\documentclass[]{fairmeta}

\usepackage{amsmath}
\usepackage{amssymb}
\usepackage{mathtools}
\usepackage{bm}
\usepackage{algorithm}
\usepackage{algpseudocode}
\usepackage{paralist}
\usepackage[most]{tcolorbox}
\usepackage[dvipsnames,table,xcdraw]{xcolor}
\usepackage{booktabs}
\usepackage{array}
\usepackage{tabularx}
\usepackage{ragged2e}
\usepackage{makecell}
\usepackage{multirow}
\usepackage{tikz}
\usepackage{pgfplots}
\usepackage{wrapfig}
\usetikzlibrary{pgfplots.groupplots, matrix}
\pgfplotsset{compat=1.18}
\usepackage{enumitem}
\usepackage{comment}
\usepackage{graphicx}
\usepackage{multirow}
\usepackage{stackengine}
\usepackage{colortbl} 
\usepackage{anyfontsize}
\usepackage{CJK}
\usepackage{subfig}
\usepackage{placeins}

\definecolor{purple}{HTML}{c994c7}
\definecolor{navyblue}{RGB}{30,130,255}
\definecolor{citecolor}{RGB}{30,130,255}
\definecolor{lightgray}{gray}{0.9}
\definecolor{blanchedalmond}{rgb}{1.0, 0.92, 0.8}
\definecolor{cerise}{rgb}{0.871, 0.192, 0.388}

\definecolor{TaskBG}{HTML}{EFE6FF}        
\definecolor{StateBG}{HTML}{F5F5F7}       
\definecolor{ExpertBG}{HTML}{EAF7EA}      
\definecolor{IWMBG}{HTML}{FDECF3}         
\definecolor{SRBG}{HTML}{E6F2FF}          

\definecolor{lastauthor}{RGB}{143, 68, 115}

\title{Beyond English: Uncovering the Multilingual Gap in Vision-Language-Action Models}

\author{Hanyang Chen}
\author{Hongliang Li}
\author{Jiarui Cao}
\author{Yang Jiang}
\author{Haonan Wen}
\author{Kaiyu Huang}
\author{Shengnan Guo}
\author{Huaiyu Wan}

\affiliation{Beijing Jiaotong University}

\abstract{
Vision-Language-Action (VLA) models have recently demonstrated promising capabilities in learning generalist robot policies from large-scale multimodal data. However, most existing VLA systems are trained and evaluated primarily with English instructions, leaving their ability to understand and execute instructions in other languages largely unexplored. While the underlying large language models often possess multilingual capabilities, it remains unclear whether these multilingual capabilities transfer to VLAs during training. In this work, we present the first systematic study of multilingual instruction following in VLA models. 
We first construct multilingual instructions by extending existing benchmarks with translations of their instructions. Using these instructions, we evaluate several representative VLA models across a range of tasks in simulation settings. Our experiments reveal a significant multilingual gap: models trained primarily on English instructions exhibit substantial performance degradation when evaluated on other languages, even when the underlying language backbone is multilingual. We provide several findings and analyses to understand the multilingual gap. Cross-lingual transfer behavior analysis shows that performance drops correlate with both instruction understanding and action execution. Representation analyses suggest that multilingual instruction-caused representation shifts may contribute to the multilingual gap.
Motivated by these findings, we further explore strategies to improve multilingual performance in VLAs. We propose a simple yet effective multilingual fine-tuning approach, Multilingual Principal Component Alignment (MPCA), which leverages Principal Component Analysis to get the principal component subspace and align projected multilingual representations, effectively reducing the multilingual performance gap. Our experiments show that MPCA can effectively improve multilingual performance in VLAs, demonstrating its potential as a practical solution for enhancing multilingual robustness in embodied agents.
}

\date{\today}
\correspondence{Hanyang Chen at \email{hanyangchen@bjtu.edu.cn}, Shengnan Guo at \email{guoshn@bjtu.edu.cn}}

\begin{document}

\maketitle

\label{sec:intro}
\begin{figure}[tbh]
    \centering
    \includegraphics[width=0.98\linewidth]{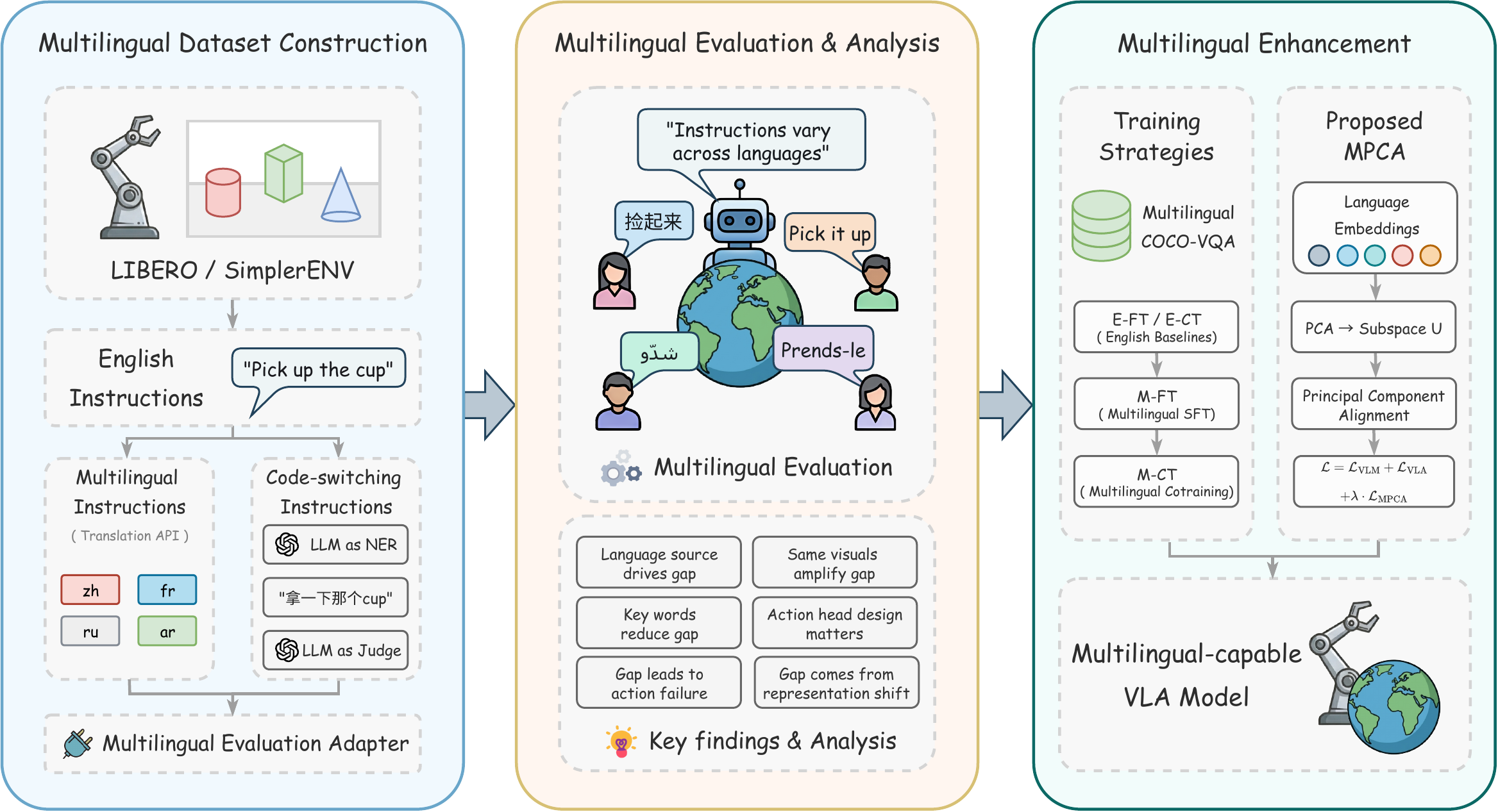}
    \caption{
        Illustration of the evaluation stages for multilingual instruction following in VLAs. We first construct multilingual instruction variants based on existing robot benchmarks, and then conduct a comprehensive evaluation and analysis to reveal the multilingual gap in current VLA systems, and finally explore strategies to improve multilingual performance in VLAs.
    }
    \label{fig:framework-figure}
\end{figure}

\section{Introduction}

Vision-Language-Action models (VLAs) have recently emerged as a promising paradigm for building generalist robot policies~\citep{team2024octo, kim2024openvla, black2024pi_0, black2025pi_, kim2025fine, cen2025rynnvla}. By integrating visual perception, natural language understanding, and action generation within a unified framework, VLAs enable robots to execute complex tasks conditioned on natural language instructions. Built upon vision-language models (VLMs), VLAs extend multimodal understanding to action generation. Recent advances in VLMs~\citep{hurst2024gpt, team2024gemini, wang2024qwen2, bai2025qwen3} have also strengthened this foundation through large-scale multimodal pretraining, further improving generalization across tasks. As a result, VLAs have become an increasingly important research direction for developing scalable and flexible embodied agents.

Despite these advances, most existing VLA research focuses primarily on improving instruction-following capability~\citep{karnik2024embodied, zhang2025vlabench, xu2025seeing} and visual grounding~\citep{fei2025libero, zhou2025libero} under English instructions. In both training and evaluation, the majority of VLA benchmarks and datasets~\citep{mees2022calvin, liu2023libero, li2023behavior, nasiriany2024robocasa, li2024evaluating, chen2025robotwin} adopt English as the default instruction language. However, this setting overlooks an important challenge in real-world deployment: robots operating in global environments must be able to understand instructions expressed in different languages. Although many VLAs rely on multilingual large language models as their backbone, it remains unclear whether these multilingual capabilities transfer to VLAs during training. In particular, the alignment process between language tokens and robot actions may implicitly bias the model toward English instructions, potentially leading to degraded performance when instructions are provided in other languages. This raises an important yet largely unexplored question: \textit {Do VLAs truly retain multilingual understanding after being aligned with robot actions?}

To address this question, a systematic evaluation of multilingual instructions following in VLAs is necessary. In this work, we study multilingual capabilities in embodied systems through three key stages, as shown in Figure~\ref{fig:framework-figure}. \textbf{(1) Multilingual dataset construction}: Evaluating multilingual generalization requires constructing instruction data across multiple languages to simulate multilingual interaction scenarios. While the underlying language models may possess multilingual knowledge, it remains uncertain whether such knowledge can be effectively transferred to the action space through language-conditioned policy learning. Therefore, we extend existing robot datasets by constructing multilingual instruction variants that allow us to directly compare model behavior across languages. \textbf{(2) Multilingual evaluation and analysis}: Beyond simply measuring task success rates, analyzing the results of VLAs under multilingual instructions can provide valuable insights into how language representations interact with action policies. In particular, we will examine how performance varies across various dimensions and investigate whether the representations remain consistent under different linguistic inputs. \textbf{(3) Multilingual performance enhancement}: Based on the insights gained from multilingual evaluation, we can further explore strategies to improve multilingual performance in VLAs.

Based on these stages, we introduce a multilingual evaluation framework for VLAs. Specifically, we construct multilingual instruction sets by extending existing robot manipulation benchmarks with instructions expressed in multiple languages. Our design focuses on two multilingual interaction settings. The first setting is the \textbf{multilingual instruction}, where instructions are translated into different languages while preserving their original semantics, enabling controlled evaluation of multilingual generalization. The second setting is the \textbf{code-switching instruction}, where instructions contain mixed-language expressions that commonly occur in multilingual communication scenarios. These settings allow us to systematically study how VLAs behave when language inputs deviate from the English-only assumption commonly used in existing benchmarks.

Using this framework, we conduct a comprehensive empirical study on several representative VLA models across two robot benchmarks, LIBERO~\citep{liu2023libero} and SimplerEnv~\citep{li2024evaluating}. Our experiments reveal a clear multilingual gap, where models trained primarily on English instructions exhibit noticeable performance degradation when evaluated on other languages. Furthermore, we analyze the multilingual performance across different base models, suites, interaction settings and action head designs, and also provide discussions from the perspective of model behavior and representations. Our findings suggest that multilingual generalization is an important yet overlooked aspect of current VLA systems, and provide insights for improving multilingual generalization for VLAs.

Based on these insights and analyses, we further explore strategies to improve multilingual performance in VLAs. We also propose a simple yet effective multilingual fine-tuning approach, Multilingual Principal Component Alignment (MPCA), which leverages Principal Component Analysis (PCA) to get the principal component subspace and align projected multilingual representations, effectively reducing the multilingual performance gap.

Our contributions are summarized as follows:
(1) We present the first systematic study of multilingual instruction following in VLA models, introducing a multilingual evaluation framework that extends existing robot benchmarks with multilingual instruction variants.
(2) We conduct comprehensive experiments across multiple VLA models, suites, and multilingual settings, revealing a significant multilingual gap in current VLA systems.
(3) We analyze the multilingual performance from both behavioral and representational perspectives, providing insights into the underlying causes of the multilingual gap and potential avenues for improvement.
(4) We propose a simple yet effective multilingual fine-tuning approach, MPCA, which effectively improves multilingual performance in VLAs.

\section{Related Work}
\label{sec:related-work}

\textbf{Vision-Language-Action Models.}
The rapid development of VLA models has been driven by large-scale pre-training and architectural innovations, yielding policies with strong English instruction-following and visual grounding capabilities.
Early approaches, such as RT-1~\citep{brohan2022rt} and Octo~\citep{team2024octo}, pioneered end-to-end transformer policies trained on massive robot datasets, establishing the foundation for generalist manipulation. Subsequent paradigms~\citep{kim2024openvla, black2024pi_0, black2025pi_, kim2025fine, cen2025rynnvla, bjorck2025gr00t} increasingly leverage pre-trained VLMs as backbones, fine-tuning them for embodied control via action discretization~\citep{kim2024openvla,kim2025fine}  or diffusion-based continuous action experts~\citep{black2024pi_0,wen2024diffusion}. To mitigate catastrophic forgetting and preserve foundation model priors during policy adaptation, recent works have explored parameter-efficient fine-tuning \citep{torne2026mem}, dual-system architectures that decouple high-level reasoning from low-level control~\citep{bjorck2025gr00t}, and co-training with external data~\citep{bu2025univla, cen2025worldvla, lian2026bayesianvla}. These architectures demonstrate remarkable zero-shot generalization in familiar settings, excelling at spatial reasoning, object grounding, and executing complex manipulation sequences conditioned on natural language. 
However, despite leveraging multilingual VLM backbones, contemporary VLA training pipelines remain overwhelmingly monolingual, which may bias the language-to-action alignment process toward English, potentially leading to degraded performance when evaluated on non-English instructions. Our work proves the existence of this multilingual gap and analyzes it in current VLA systems.

\textbf{Evaluation Benchmarks for VLAs.}
Evaluation benchmarks for VLA models have progressively expanded from static task completion metrics to systematic robustness analysis across visual, kinematic, and environmental dimensions.
Previous benchmarks, such as CALVIN~\citep{mees2022calvin} and LIBERO~\citep{liu2023libero}, established standardized environments for language-conditioned manipulation. To expose hidden brittleness, recent benchmarks~\citep{fei2025libero, zhou2025libero, chen2025robotwin, wang2025vlatest} have automated perturbation generation and introduced multi-dimensional evaluation protocols. These benchmarks systematically evaluate model resilience to camera viewpoint shifts, object layout variations, lighting changes, robot initial state perturbations, and sensor noise, revealing that contemporary VLAs are highly sensitive to spatial and visual distribution shifts. However, regarding language robustness, existing evaluations~\citep{fei2025libero, wang2025vlatest} typically restrict perturbations to English paraphrasing or instruction complexity variations. While these studies confirm that VLAs can handle semantic variations within English, they operate under the implicit assumption that language understanding is language-invariant once trained. However, this assumption overlooks the real-world needs for multilingual instruction following. This omission creates a critical gap in embodied evaluation: without multilingual evaluation, performance degradation across languages cannot be quantified, limiting real-world deployment. Our work addresses this gap by systematically evaluating multilingual instruction following in VLAs, revealing a significant multilingual gap and providing insights into the underlying reasons and potential remedies.

\section{Multilingual Evaluation Framework for VLAs}
\label{sec:method}

In this section, we introduce our multilingual evaluation framework for VLAs, which consists of two key components: (1) a multilingual instruction construction pipeline that generates multilingual instruction variants for existing robot benchmarks; and (2) a multilingual evaluation adapter that enables seamless integration of multilingual instructions into the evaluation process. This framework allows us to systematically assess the multilingual generalization capabilities of VLA models under controlled and consistent conditions.

\subsection{Multilingual Instruction Construction}

\subsubsection{Multilingual Settings Design}

\textbf{Expanding the scope of languages.}
Following most of the prior works in multilingual LLMs and VLMs~\citep{thellmann2024towards, yong2025state, luo2026lost}, we select Chinese, French, Russian, and Arabic as the target languages for evaluation, covering high-resource languages (English and Chinese) and low-resource languages (French, Russian, and Arabic) in different language families and scripts. 

\textbf{Introducing diverse multilingual interaction settings.}
We consider two multilingual interaction settings in this paper: (1) multilingual instructions, where instructions are directly translated into different languages while preserving their original semantics; and (2) code-switching instructions, where instructions contain mixed-language expressions that commonly occur in multilingual communication scenarios. These settings allow us to systematically study how VLAs behave when language inputs deviate from the English-only assumption commonly used in existing benchmarks.

\subsubsection{Instruction Generation Pipeline}

To construct multilingual instruction variants, we start from the original English instructions provided in existing benchmarks. For the multilingual instruction setting, we use a standard machine translation system, i.e., Cloud Translation API, to translate the English instructions into each target language. For the code-switching instruction setting, we use LLMs to first perform named entity recognition, identifying key verbs and nouns in both the original English instructions and their translated versions. Based on these alignments, the LLMs then substitute the corresponding key phrases in the translated instructions with those from the original English instructions, resulting in code-switched instructions. To ensure the quality of the generated instructions, we employ LLM-as-a-judge to evaluate the semantic consistency between the original English instructions and the code-switching instructions. This pipeline allows us to systematically generate multilingual instruction variants while controlling for semantic equivalence, enabling a rigorous evaluation of multilingual generalization in VLA models.

\subsection{Multilingual Evaluation Adapter}

To achieve a simple and fair evaluation of multilingual performance across different VLA models, we provide a multilingual evaluation adapter that can replace the original language instruction with only one line of code change. This adapter takes the original instruction as input and outputs the corresponding multilingual instruction variant based on the specified language setting. By integrating this adapter into the evaluation pipeline, we can seamlessly switch between different language inputs without modifying any other components of the model or environment. This design ensures that all models are evaluated under consistent conditions, allowing for a direct comparison of multilingual performance while maintaining the integrity of the original evaluation protocol. We provide instruction variants for LIBERO~\citep{liu2023libero} and SimplerEnv~\citep{li2024evaluating}, and the adapter can automatically process new instructions and languages to generate corresponding variants.

\subsection{Evaluation Protocol}

We follow the original evaluation protocols of the respective benchmarks for each task, measuring task success rates under different language settings. To quantify the multilingual performance gap, we report relative performance with respect to the original English instructions, allowing us to directly compare the impact of language changes on model performance:
\begin{equation}
    \text{Relative Performance} = (\text{Success Rate}_{\text{Multilingual}} - \text{Success Rate}_{\text{English}}) \times 100\%.
\end{equation}

\section{Experiments}
\label{sec:experiment}

In this section, we first introduce the experiment setup in Section~\ref{subsec:experiment-setup}, including the environments and baselines used for evaluation. We then present the main results in Section~\ref{subsec:main-results}, where we analyze the multilingual performance of different VLA models across various languages and multilingual settings. We further investigate how the multilingual gap relates to model behavior in Section~\ref{subsubsec:model-behavior-analysis}, and analyze the underlying reasons behind the multilingual gap in Section~\ref{subsubsec:representation-shift-analysis}. Finally, we explore potential strategies to improve the multilingual capability of VLA models in Section~\ref{subsec:multilingual-capability-improvement}.

\subsection{Experiment Setup}
\label{subsec:experiment-setup}

\textbf{Environments.} We evaluate various VLA models on two simple but representative simulation benchmarks: LIBERO~\citep{liu2023libero} and SimplerEnv~\citep{li2024evaluating}. These environments cover a range of tasks with varying complexity, such as object rearrangement and tool use. Each environment provides a standardized evaluation protocol and metrics for assessing task success, enabling a systematic comparison of multilingual performance across diverse settings.

\textbf{Baselines.} We compare several VLA models that have been trained primarily on English instructions, including~$\pi_{0.5}$~\citep{intelligence2025pi_}, OpenVLA-OFT~\citep{kim2025fine}, and ABot-M0~\citep{yang2026abot}. Multiple Qwen-VL-based VLAs with different action head designs~\citep{community2026starvla} are also included to evaluate the impact of multilingual VLM backbones and architectural choices on multilingual performance.
In addition, we include a world-model-based policy, Cosmos Policy~\citep{kim2026cosmos}, to investigate existing world-model-based approaches in multilingual settings. We provide detailed information about these models in the Appendix~\ref{subsec:model-details}.

\subsection{Main Results}
\label{subsec:main-results}

\begin{table}[t]
\caption{Multilingual performance on LIBERO across four suites, with the average performance across suites shown in the last group. Relative performance is shown with respect to the original English instructions. $\varnothing$ denotes evaluated without any instructions. The Avg. denotes the average performance under each language across the four suites. Percentage sign is omitted for better readability.}
\centering
\small
\setlength{\tabcolsep}{3.5pt}
\renewcommand{\arraystretch}{1.15}
\resizebox{\textwidth}{!}{
\begin{tabular}{l|c|ccccc|ccccc|ccccc|ccccc|ccccc}
\toprule

\multicolumn{2}{c|}{\multirow{2}{*}{Models}}
& \multicolumn{5}{c|}{Long}
& \multicolumn{5}{c|}{Goal}
& \multicolumn{5}{c|}{Object}
& \multicolumn{5}{c|}{Spatial}
& \multicolumn{5}{c}{Avg.} \\
\cmidrule(lr){3-7} \cmidrule(lr){8-12} \cmidrule(lr){13-17} \cmidrule(lr){18-22} \cmidrule(lr){23-27}

\multicolumn{2}{c|}{}
& zh & fr & ru & ar & $\varnothing$
& zh & fr & ru & ar & $\varnothing$
& zh & fr & ru & ar & $\varnothing$
& zh & fr & ru & ar & $\varnothing$
& zh & fr & ru & ar & $\varnothing$ \\

\midrule

\multicolumn{2}{c|}{OpenVLA-OFT}
& -12.5 & -12.5 & -9.5 & -13.5 & -8.5
& -89.0 & -89.0 & -89.0 & -89.0 & -89.0
& -2.0 & -1.0 & -1.5 & -2.0 & -1.0
& -11.0 & -12.0 & -11.0 & -14.5 & -11.5
& -28.6 & -28.6 & -27.8 & -29.8 & -27.5 \\

\multicolumn{2}{c|}{$\pi_{0.5}$}
& -20.0 & -17.0 & -23.5 & -21.0 & -23.0
& -78.0 & -84.5 & -88.5 & -85.5 & -85.0
& -32.5 & -28.0 & -31.5 & -33.5 & -34.5
& -28.0 & -22.5 & -23.5 & -25.5 & -31.5
& -39.6 & -38.0 & -41.8 & -41.4 & -43.5 \\

\multicolumn{2}{c|}{ABot-M0}
& -5.5 & -5.0 & -14.0 & -15.5 & -18.0
& -33.5 & -88.0 & -89.5 & -87.0 & -87.5
& -4.5 & -19.0 & -11.5 & -8.5 & -8.5
& -3.5 & -9.0 & -7.5 & -15.5 & -13.5
& -11.8 & -30.3 & -30.6 & -31.6 & -31.9 \\

\multicolumn{2}{c|}{Cosmos Policy}
& -19.5 & -3.5 & -14.5 & -20.5 & -22.0
& -89.0 & -46.5 & -81.5 & -87.5 & -87.5
& -38.0 & -2.0 & -28.0 & -38.0 & -41.0
& -39.0 & -16.0 & -31.0 & -40.0 & -41.0
& -46.4 & -17.0 & -38.8 & -46.5 & -47.9 \\

\midrule
{\footnotesize\multirow{4}{*}{\rotatebox{90}{\textbf{Qwen2.5-VL}}}}
& OFT
& -7.0 & -22.0 & -26.0 & -35.5 & -32.9
& -18.0 & -71.5 & -71.5 & -90.0 & -92.0
& -19.0 & -8.5 & -34.5 & -42.0 & -23.4
& -27.0 & -26.5 & -49.0 & -69.0 & -25.0
& -17.8 & -32.1 & -45.3 & -59.1 & -43.3 \\

& FAST
& -22.5 & -51.0 & -55.0 & -52.0 & -52.5
& -20.0 & -83.5 & -80.0 & -83.5 & -83.0
& -30.0 & -46.0 & -47.0 & -48.5 & -46.5
& -31.0 & -82.0 & -81.5 & -76.5 & -84.5
& -25.9 & -65.6 & -65.9 & -65.1 & -66.6 \\

& GR00T
& 3.0 & -14.5 & -21.0 & -26.5 & -28.5
& -17.5 & -69.5 & -64.5 & -80.5 & -93.0
& -10.0 & -8.5 & -22.0 & -36.0 & -22.0
& -13.5 & -11.5 & -22.5 & -23.5 & -33.0
& -9.5 & -26.0 & -32.5 & -41.6 & -44.1 \\

& $\pi$
& -16.5 & -22.5 & -31.0 & -26.5 & -29.0
& -57.5 & -76.5 & -79.0 & -83.0 & -83.5
& -22.0 & -27.0 & -27.5 & -35.5 & -39.0
& -34.5 & -27.0 & -39.0 & -37.0 & -35.5
& -32.6 & -38.3 & -44.1 & -45.5 & -46.8 \\

\midrule
{\footnotesize\multirow{4}{*}{\rotatebox{90}{\textbf{Qwen3-VL}}}}
& OFT
& 3.0 & -23.0 & -33.0 & -33.0 & -46.0
& -3.0 & -78.0 & -91.5 & -79.0 & -90.5
& -13.5 & -18.0 & -14.5 & -24.5 & -57.5
& -15.5 & -37.5 & -56.0 & -52.0 & -58.5
& -7.3 & -39.1 & -48.8 & -47.1 & -63.1 \\

& FAST
& -25.5 & -65.5 & -70.0 & -69.0 & -63.5
& -32.0 & -80.0 & -82.5 & -78.5 & -81.0
& -26.5 & -63.0 & -68.5 & -64.5 & -61.0
& -34.5 & -74.5 & -84.5 & -82.5 & -76.5
& -29.6 & -70.8 & -76.4 & -73.6 & -70.5 \\

& GR00T
& 1.0 & -19.5 & -23.0 & -26.0 & -34.0
& -2.0 & -87.0 & -84.0 & -75.0 & -92.5
& -9.0 & -14.0 & -34.0 & -25.0 & -36.0
& -14.5 & -11.5 & -14.0 & -18.0 & -29.5
& -6.1 & -33.0 & -38.8 & -36.0 & -48.0 \\

& $\pi$
& -6.0 & -22.5 & -18.5 & -26.5 & -36.5
& -10.5 & -90.0 & -87.0 & -81.5 & -91.0
& -10.0 & -33.5 & -47.5 & -42.5 & -42.0
& -22.0 & -17.5 & -25.5 & -36.5 & -52.0
& -12.1 & -40.9 & -44.6 & -46.8 & -55.4 \\

\bottomrule
\end{tabular}
}
\label{tab:multilingual-results}
\end{table}
\begin{table}[t]
\caption{Multilingual performance on SimperEnv across four tasks, with the average performance across tasks shown in the last group. Relative results are shown with respect to the original English instructions. $\varnothing$ denotes evaluated without any instructions. The Avg. denotes the average performance under each language across the four suites. Percentage sign is omitted for better readability.}
\centering
\small
\setlength{\tabcolsep}{3.5pt}
\renewcommand{\arraystretch}{1.15}
\resizebox{\textwidth}{!}{
\begin{tabular}{l|c|ccccc|ccccc|ccccc|ccccc|ccccc}
\toprule

\multicolumn{2}{c|}{\multirow{2}{*}{Models}}
& \multicolumn{5}{c|}{\makecell{Put Spoon \\ on Towel}}
& \multicolumn{5}{c|}{\makecell{Put Carrot \\ on Plate}}
& \multicolumn{5}{c|}{\makecell{Stack Green Block \\ on Yellow Block}}
& \multicolumn{5}{c|}{\makecell{Put Eggplant \\ in Yellow Basket}}
& \multicolumn{5}{c}{Avg.} \\
\cmidrule(lr){3-7} \cmidrule(lr){8-12} \cmidrule(lr){13-17} \cmidrule(lr){18-22} \cmidrule(lr){23-27}

\multicolumn{2}{c|}{}
& zh & fr & ru & ar & $\varnothing$
& zh & fr & ru & ar & $\varnothing$
& zh & fr & ru & ar & $\varnothing$
& zh & fr & ru & ar & $\varnothing$
& zh & fr & ru & ar & $\varnothing$ \\

\midrule
{\footnotesize\multirow{4}{*}{\rotatebox{90}{\textbf{Qwen2.5-VL}}}}
& OFT
& -33.3 & -25.0 & -25.0 & -20.8 & -37.5
& -20.8 & -16.7 & -16.7 & -20.8 & -33.3
& -8.3 & -8.3 & -8.3 & -8.3 & -8.3
& -58.3 & -50.0 & -50.0 & -58.3 & -87.5
& -30.2 & -25.0 & -25.0 & -27.1 & -41.7 \\

& FAST
& -29.2 & -12.5 & -37.5 & -29.2 & -79.2
& -4.2 & -12.5 & -16.7 & -50.0 & -50.0
& -12.5 & -12.5 & -29.2 & -29.2 & -37.5
& -37.5 & 0.0 & -54.2 & 12.5 & -83.3
& -20.8 & -9.4 & -34.4 & -24.0 & -62.5 \\

& GR00T
& 0.0 & -1.0 & -72.9 & -76.0 & -84.4
& -1.0 & -13.5 & -29.2 & -54.2 & -51.0
& -8.3 & -23.6 & -37.5 & -41.7 & -41.7
& -10.4 & -11.5 & -49.0 & 8.3 & -66.7
& -4.9 & -12.4 & -47.1 & -40.9 & -60.9 \\

& $\pi$
& 12.5 & 4.2 & -62.5 & -79.2 & -79.2
& -16.7 & -33.3 & -25.0 & -58.3 & -79.2
& -12.5 & -16.7 & -12.5 & -29.2 & -29.2
& 20.8 & 8.3 & 25.0 & -50.0 & -62.5
& 1.0 & -9.4 & -18.8 & -54.2 & -62.5 \\

\bottomrule
\end{tabular}
}
\label{tab:multilingual-results-simpler-env}
\end{table}

\begin{table}[t]
\caption{Code-switching performance on LIBERO across four suites, with the average performance across suites shown in the last group. Relative results are shown with respect to the original English instructions. The Avg. denotes the average performance under each language across the four suites. Percentage sign is omitted for better readability.}
\centering
\small
\setlength{\tabcolsep}{3.5pt}
\renewcommand{\arraystretch}{1.15}
\resizebox{1.0\textwidth}{!}{
\begin{tabular}{l|c|cccc|cccc|cccc|cccc|cccc}
\toprule

\multicolumn{2}{c|}{\multirow{2}{*}{Models}}
& \multicolumn{4}{c|}{Long}
& \multicolumn{4}{c|}{Goal}
& \multicolumn{4}{c|}{Object}
& \multicolumn{4}{c|}{Spatial}
& \multicolumn{4}{c}{Avg.} \\
\cmidrule(lr){3-6} \cmidrule(lr){7-10} \cmidrule(lr){11-14} \cmidrule(lr){15-18} \cmidrule(lr){19-22}
\multicolumn{2}{c|}{}
& zh & fr & ru & ar
& zh & fr & ru & ar
& zh & fr & ru & ar
& zh & fr & ru & ar
& zh & fr & ru & ar \\

\midrule

\multicolumn{2}{c|}{OpenVLA-OFT}
& -14.5 & -10.5 & -10.0 & -4.0
& -73.0 & -45.0 & -53.5 & -50.5
& -2.5 & 0.0 & -1.0 & -1.5
& -11.0 & -1.5 & -5.0 & -12.0
& -25.25 & -14.25 & -17.38 & -17.00 \\

\multicolumn{2}{c|}{$\pi_{0.5}$}
& -4.0 & -2.5 & -8.5 & -12.0
& -17.0 & -28.0 & -32.0 & -33.5
& -13.0 & -1.5 & -6.5 & -18.5
& -21.0 & -16.5 & -20.0 & -23.5
& -13.75 & -12.12 & -16.75 & -21.88 \\

\multicolumn{2}{c|}{ABot-M0}
& -0.5 & -1.5 & -13.0 & -12.5
& -22.0 & -27.0 & -37.5 & -40.0
& -6.5 & -1.5 & -1.0 & -6.0
& 1.0 & -10.5 & -6.0 & -6.5
& -7.00 & -10.13 & -14.38 & -16.25 \\

\multicolumn{2}{c|}{Cosmos Policy}
& -15.5 & -6.5 & -3.0 & -11.0
& -50.0 & -21.0 & -36.0 & -61.0
& -25.0 & -2.5 & -5.0 & -20.5
& -40.0 & -9.0 & -21.5 & -30.0
& -32.63 & -9.75 & -16.38 & -30.63 \\

\midrule
{\footnotesize\multirow{4}{*}{\rotatebox{90}{\textbf{Qwen2.5-VL}}}}
& OFT
& -0.5 & -15.0 & -21.5 & -20.5
& -11.5 & -20.5 & -32.0 & -58.0
& -11.5 & -3.0 & -1.5 & -19.0
& -28.0 & -22.5 & -33.5 & -51.5
& -12.88 & -15.25 & -22.13 & -37.25 \\

& FAST
& -20.0 & -47.5 & -46.5 & -43.0
& -20.5 & -63.0 & -51.5 & -53.0
& -30.0 & -43.0 & -37.5 & -43.5
& -30.5 & -74.0 & -30.5 & -71.0
& -25.25 & -56.88 & -41.50 & -52.63 \\

& GR00T
& -6.5 & -9.0 & -15.0 & -14.0
& -17.0 & -26.0 & -27.0 & -35.5
& -9.0 & -4.5 & -12.0 & -19.0
& -18.5 & -13.5 & -20.5 & -20.0
& -12.75 & -13.25 & -18.63 & -22.13 \\

& $\pi$
& -15.0 & -14.5 & -24.5 & -24.0
& -42.5 & -55.0 & -49.5 & -55.5
& -13.5 & -13.5 & -9.5 & -23.0
& -35.5 & -27.0 & -36.0 & -37.5
& -26.63 & -27.50 & -29.88 & -35.00 \\

\midrule
{\footnotesize\multirow{4}{*}{\rotatebox{90}{\textbf{Qwen3-VL}}}}
& OFT
& 1.0 & -18.5 & -7.0 & -16.5
& -3.0 & -19.5 & -34.0 & -31.0
& -10.5 & -0.5 & -5.0 & -10.5
& -14.5 & -30.5 & -43.0 & -52.0
& -6.75 & -17.25 & -22.25 & -27.50 \\

& FAST
& -26.0 & -63.5 & -59.5 & -60.0
& -23.0 & -55.0 & -52.0 & -46.5
& -37.0 & -43.5 & -44.5 & -55.0
& -37.0 & -62.0 & -79.0 & -77.0
& -30.75 & -56.00 & -58.75 & -59.63 \\

& GR00T
& 1.5 & -14.5 & -11.5 & -20.5
& 1.0 & -30.5 & -35.0 & -30.0
& -8.5 & -0.5 & -0.5 & -8.0
& -14.5 & -8.0 & -16.0 & -12.5
& -5.13 & -13.38 & -15.75 & -17.75 \\

& $\pi$
& -4.0 & -11.0 & -14.5 & -14.5
& -1.0 & -38.0 & -35.5 & -45.0
& -11.0 & -14.0 & -6.5 & -28.5
& -20.5 & -9.0 & -20.5 & -31.0
& -9.13 & -18.00 & -19.25 & -29.75 \\

\bottomrule
\end{tabular}
}
\label{tab:code-switching-results}
\end{table}

\begin{table}[t]
\caption{Code-switching performance on SimplerEnv across four tasks, with the average performance across tasks shown in the last group. Relative results are shown with respect to the original English instructions. The Avg. denotes the average performance under each language across the four suites. Percentage sign is omitted for better readability.}
\centering
\small
\setlength{\tabcolsep}{3.5pt}
\renewcommand{\arraystretch}{1.15}
\resizebox{1.0\textwidth}{!}{
\begin{tabular}{l|c|cccc|cccc|cccc|cccc|cccc}
\toprule

\multicolumn{2}{c|}{\multirow{2}{*}{Models}}
& \multicolumn{4}{c|}{\makecell{Put Spoon \\ on Towel}}
& \multicolumn{4}{c|}{\makecell{Put Carrot \\ on Plate}}
& \multicolumn{4}{c|}{\makecell{Stack Green Block \\ on Yellow Block}}
& \multicolumn{4}{c|}{\makecell{Put Eggplant \\ in Yellow Basket}}
& \multicolumn{4}{c}{Avg.} \\
\cmidrule(lr){3-6} \cmidrule(lr){7-10} \cmidrule(lr){11-14} \cmidrule(lr){15-18} \cmidrule(lr){19-22}

\multicolumn{2}{c|}{}
& zh & fr & ru & ar
& zh & fr & ru & ar
& zh & fr & ru & ar
& zh & fr & ru & ar
& zh & fr & ru & ar \\

\midrule
{\footnotesize\multirow{4}{*}{\rotatebox{90}{\textbf{Qwen2.5-VL}}}}
& OFT
& -33.3 & -25.0 & -25.0 & -37.5
& -20.8 & -12.5 & -25.0 & -20.8
& -4.2 & -8.3 & -4.2 & -8.3
& -62.5 & -41.7 & -70.8 & -83.3
& -30.2 & -21.9 & -31.3 & -34.4 \\

& FAST
& 8.3 & -16.7 & 4.2 & -58.3
& -12.5 & -4.2 & -12.5 & -12.5
& -20.8 & -16.7 & -16.7 & -37.5
& 12.5 & -16.7 & 12.5 & 4.2
& -3.1 & -13.5 & -3.1 & -26.0 \\

& GR00T
& -9.4 & -13.5 & 3.1 & -51.0
& -1.0 & -17.7 & -13.5 & -17.7
& -25.0 & -37.5 & -33.3 & -41.7
& 4.2 & -4.2 & 12.5 & -4.2
& -7.8 & -18.2 & -7.8 & -28.6 \\

& $\pi$
& 4.2 & 6.3 & 8.3 & -45.8
& -12.5 & -33.3 & -16.7 & -29.2
& -16.7 & -25.0 & -4.2 & -25.0
& 4.2 & 0.0 & 12.5 & 0.0
& -9.4 & -13.0 & 0.0 & -25.0 \\

\bottomrule
\end{tabular}
}
\label{tab:code-switching-results-simpler-env}
\end{table}

We provide the relative performance of all models across different languages and multilingual settings in Table~\ref{tab:multilingual-results}, Table~\ref{tab:multilingual-results-simpler-env}, Table~\ref{tab:code-switching-results} and Table~\ref{tab:code-switching-results-simpler-env}. Results with no instruction are also included as a reference to show the performance of models without any language input. We analyze the results from three perspectives and provide our findings for each perspective. Additional results are included in the Appendix~\ref{subsec:extended-main-results}.

\textbf{Finding 1: Multilingual gaps are driven by the language source of the base VLM.} 
We analyze the relative performance of each language compared to English for each model, and we observe a consistent performance gap between English and non-English instructions across almost all models in both LIBERO and SimplerEnv environments. We first divide the models into two groups based on their base VLMs: (1) $\pi_{0.5}$, OpenVLA-OFT are based on PaliGemma~\citep{steiner2024paligemma} and Prismatic VLM~\citep{karamcheti2024prismatic}, and Cosmos Policy is a world-model-based policy, which are all trained on English-centric data; (2) ABot-M0 and Qwen-VL-based VLAs are based on Qwen-VL~\citep{bai2025qwen2, bai2025qwen3}, which is a multilingual VLM trained on English, Chinese, and additional multilingual data. The first group of models shows significant performance drops on non-English instructions, which are close to the performance of models with no instructions. This suggests that these models struggle to understand and follow instructions in non-English languages, likely due to the lack of multilingual training data in their base VLMs. The second group of models shows better multilingual performance, especially on Chinese instructions. This suggests that the multilingual training data in Qwen-VL has enabled ABot-M0 and other Qwen-VL-based VLAs to generalize well to Chinese instructions, despite the absence of explicit training on Chinese instructions. These two groups reveal a significant insight that the multilingual training data in the base VLM can significantly impact the multilingual performance of the resulting VLA models, and that models trained on English-centric data may struggle to generalize to instructions in other languages.

\textbf{Finding 2: Same visual input zooms in the multilingual gap.} We observe that the multilingual gap is more pronounced in the LIBERO-Goal environment compared to other environments. This is because LIBERO-Goal contains almost the same visual input across different tasks, which makes the language input more critical for task success. Therefore, we consider that the language understanding should be a key factor that contributes to the performance gap across different languages, which is also mentioned in the previous work~\citep{lian2026langforce}.

\textbf{Finding 3: Key words in instructions can help reduce the multilingual gap.} We compare the performance of models in the multilingual instruction setting and the code-switching instruction setting, where instructions contain several key verbs or nouns in English. We observe that the performance drop in the code-switching setting is generally smaller than that in the multilingual instruction setting across all models and environments. This suggests that the presence of key words in the instructions can provide helpful cues for the models to better understand and follow the instructions, even when the overall instruction is in a different language. This finding highlights that the keywords in instructions are crucial for the language understanding of VLA models.

\textbf{Finding 4: The action head plays an essential role in multilingual generalization.} We compare the multilingual performance of models with different action head designs, including OFT-style, FAST-style, $\pi$-style, and GR00T-style action heads. With the same base VLM, we observe that models with GR00T-style and $\pi$-style action heads exhibit better multilingual performance compared to models with other action head designs. In contrast, models with FAST-style action heads show worse multilingual performance. This is because multilingual instructions would shift the distribution of the representations from the base VLM, which is supported by the visualization in Section~\ref{subsec:multilingual-capability-improvement}, and the design of action heads can significantly impact the model's ability to adapt to this distribution shift. GR00T-style and $\pi$-style action heads generate actions through a diffusion transformer, which can retain the semantic information from the base VLM and better adapt to the distribution shift caused by multilingual instructions. While FAST-style action heads directly use FAST tokenizers to process the output of the base VLM, which may lead to a loss of semantic information and worse multilingual performance. We also observe OFT-style action heads sometimes perform in between. We think that the MLP module in the OFT-style action head cannot effectively adapt to the distribution shift caused by multilingual instructions. Overall, this finding suggests that the design of action heads can play a crucial role in the multilingual generalization of VLA models, and that GR00T-style and $\pi$-style action heads with diffusion transformers are more effective in handling multilingual instructions.

\subsection{Analysis}
\label{subsec:analysis}

We further analyze the multilingual gap from two perspectives: (1) how the multilingual gap reflects model behavior, and (2) where the multilingual gap comes from.

\subsubsection{Model Behavior Reflects the Multilingual Gap}
\label{subsubsec:model-behavior-analysis}

\begin{figure}
    \centering
    \subfloat[Behavior of Qwen3-VL-$\pi$ with the instruction in French: "turn on the stove"]{
        \includegraphics[width=0.99\linewidth]{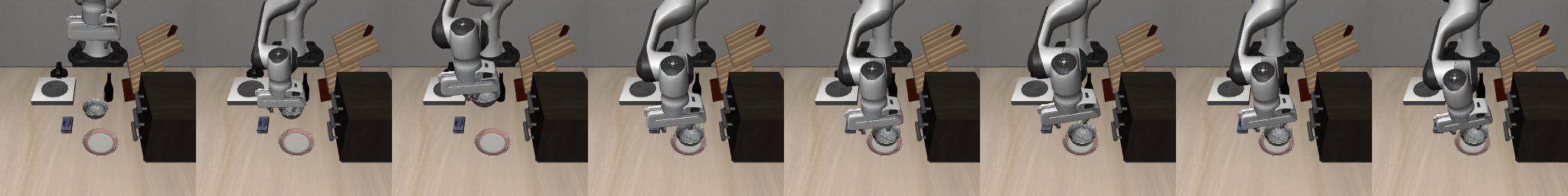}
        \label{fig:instruction-confusion}
    }
    \\
    \subfloat[Behavior of Qwen2.5-FAST with the instruction in Chinese: "pick up the cream cheese and place it in the basket"]{
        \includegraphics[width=0.99\linewidth]{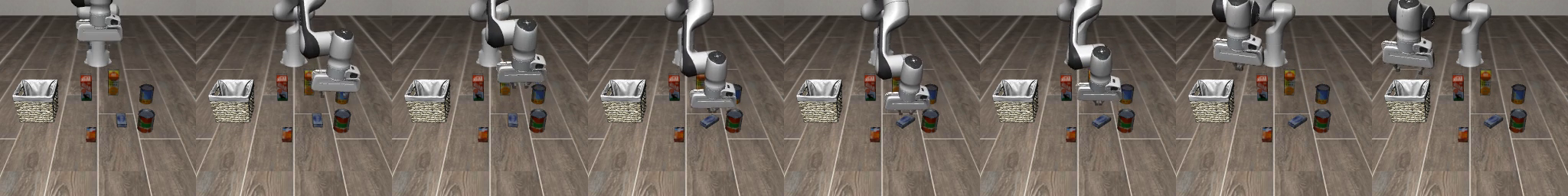}
        \label{fig:instruction-action-failure}
    }
    \caption{
        Behavioral analysis of the failure cases under multilingual instructions. Case (a) shows a failure case on LIBERO-Goal, where the model confuses the task of "turn on the stove" with "put the bowl on the plate". Case (b) shows a failure case on LIBERO-object, where the model recognizes the instruction but fails to execute the correct action.
    }
    \label{fig:failure-cases}
\end{figure}

In this section, we are going to analyze the relationship between the multilingual gap and the model behavior. We analyze multiple failure cases under multilingual instructions and find two key failure modes. The first failure mode is that the model fails to understand the instruction. This failure mode will lead to the model completely misinterpreting the instruction and generating confusing actions. Meanwhile, similar visual input may further exacerbate the confusion, which is the case in LIBERO-Goal. We provide an example in Figure~\ref{fig:instruction-confusion}, where Qwen3-VL-$\pi$ fails to understand the instruction "turn on the stove" in French, and it completely misinterprets the instruction "put the bowl on the plate" due to the similar visual input. 
The second failure mode is that the model recognizes the instruction but fails to execute the correct action. This failure mode suggests that the model may have some understanding of the instruction but still struggles to generate the correct action. We provide an example in Figure~\ref{fig:instruction-action-failure}, where Qwen2.5-FAST recognizes the instruction "pick up the cream cheese and place it in the basket" in Chinese and generates an action that is related to picking up the cream cheese, but it fails to execute the correct action of picking up the cream cheese and placing it in the basket. These two failure modes suggest that the multilingual gap can be reflected in the language understanding and action generation of VLA models. We also provide more failure cases in the Appendix~\ref{subsec:extended-model-behavior-analysis}, which further support our analysis.

\subsubsection{Representation Shift Leads to Multilingual Gap}
\label{subsubsec:representation-shift-analysis}

\begin{wrapfigure}{r}{0.4\textwidth}
    \centering
    \vspace*{-13pt} 
    \includegraphics[width=1.0\linewidth]{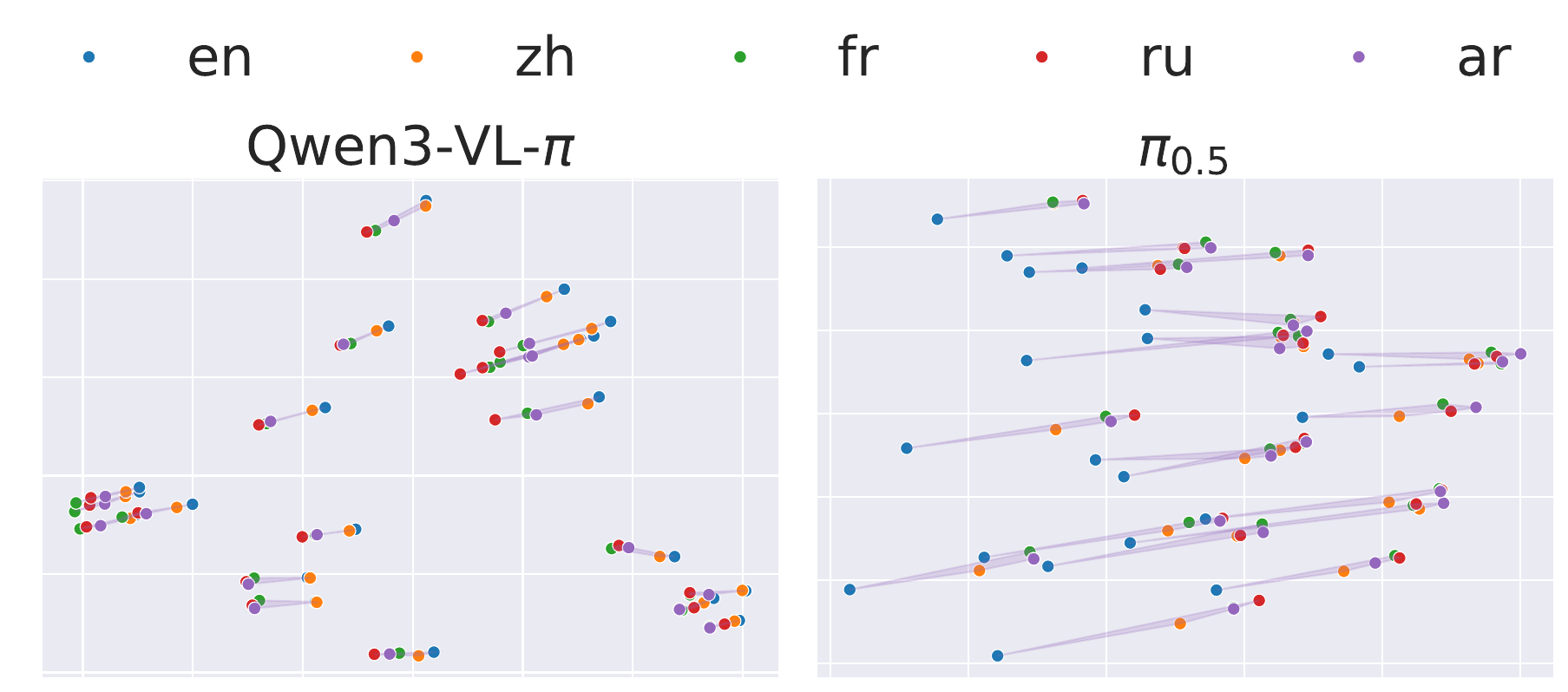}
        \caption{
            Average pooled embeddings from the middle layer of Qwen3-VL-$\pi$ and $\pi_{0.5}$. The same instruction in different languages are connected with polygon lines.
        }
        \label{fig:embedding-comparison}
\end{wrapfigure}

In this section, we are going to further analyze the reason behind the multilingual gap. As shown in Figure~\ref{fig:embedding-comparison}, we provide the visualization of the average pooled embedding from the middle layer of Qwen3-VL-$\pi$ and $\pi_{0.5}$. The same instruction in different languages is connected with polygon lines. We observe that the embeddings of English and Chinese instructions are relatively close in Qwen3-VL-$\pi$, while the embeddings of other languages are far away from English and Chinese. In $\pi_{0.5}$, the embeddings of all non-English instructions are far away from English. Obviously, the similarity of the principal components among different languages correlates well with the performance gap, discussed in Section~\ref{subsec:main-results}. The better multilingual performance of Qwen3-VL-$\pi$ can be attributed to the better cross-lingual alignment between English and Chinese, which may help the model better generalize to other languages as well. We also provide multiple suite-wise visualizations for different models in the Appendix~\ref{subsec:extended-representation-shift-analysis}, which further support our analysis.

\subsection{Multilingual performance enhancement}
\label{subsec:multilingual-capability-improvement}

\begin{table}[t]
\caption{Multilingual performance of different training strategies on LIBERO across four suites. Absolute performance is shown. \textbf{Bold} and \underline{underline} denote the best and second-best performance under each language, respectively. The Avg. denotes the average performance under each language across the four suites. Percentage sign is omitted for better readability.}
\centering
\small
\setlength{\tabcolsep}{3.5pt}
\renewcommand{\arraystretch}{1.15}
\resizebox{\textwidth}{!}
{
\begin{tabular}{c|ccccc|ccccc|ccccc|ccccc|ccccc}
\toprule

\multirow{2}{*}{Models}
& \multicolumn{5}{c|}{Long}
& \multicolumn{5}{c|}{Goal}
& \multicolumn{5}{c|}{Object}
& \multicolumn{5}{c|}{Spatial}
& \multicolumn{5}{c}{Avg.} \\
\cmidrule(lr){2-6} \cmidrule(lr){7-11} \cmidrule(lr){12-16} \cmidrule(lr){17-21} \cmidrule(lr){22-26} 

\multicolumn{1}{c|}{}
& en & zh & fr & ru & ar
& en & zh & fr & ru & ar
& en & zh & fr & ru & ar
& en & zh & fr & ru & ar
& en & zh & fr & ru & ar \\

\midrule
GR00T
& 92.5 & \textbf{93.5} & 73.0 & 69.5 & 66.5
& \underline{97.5} & \underline{95.5} & 10.5 & 13.5 & 22.5
& 97.0 & \textbf{88.0} & 83.0 & 63.0 & 72.0
& 94.0 & 79.5 & 82.5 & 80.0 & 76.0
& 95.3 & 89.1 & 62.3 & 56.5 & 59.3 \\

\midrule

E-FT
& \textbf{94.5} & \underline{91.5} & 62.0 & 69.0 & 64.0 
& \textbf{98.0} & 92.0 & 12.5 & 6.0 & 27.0 
& \textbf{100.0} & 80.5 & 91.0 & 72.5 & 69.0
& 92.5 & 73.0 & 54.5 & 46.5 & 51.0
& \textbf{96.3} & 84.3 & 55.0 & 48.5 & 52.8 \\

M-FT
& \underline{93.5} & 90.0 & \underline{78.5} & 77.5 & \textbf{84.5} 
& \textbf{98.0} & \textbf{97.5} & 23.5 & 23.0 & 43.0 
& \underline{98.5} & \underline{87.5} & 74.5 & 59.0 & 59.5 
& 91.5 & 75.5 & 84.0 & 69.0 & 64.0
& \underline{95.4} & 87.6 & 65.1 & 57.1 & 62.8 \\

\midrule

E-CT
& 88.5 & 86.0 & 78.0 & 72.5 & 75.0 
& 96.5 & 91.5 & 23.0 & 14.5 & 25.0 
& 98.0 & 84.5 & 82.5 & 59.5 & 65.5 
& 93.0 & \underline{85.5} & 89.5 & 71.5 & 64.0
& 94.0 & 86.9 & 68.3 & 54.5 & 57.4 \\

M-CT
& 80.0 & 82.5 & \underline{78.5} & \underline{78.0} & \underline{83.5} 
& 96.5 & 94.5 & \textbf{81.5} & \underline{67.0} & \underline{86.5} 
& \underline{98.5} & 87.0 & \underline{96.0} & \underline{94.0} & \underline{94.5} 
& \underline{95.5} & \textbf{89.0} & \underline{93.5} & \textbf{93.5} & \textbf{93.0}
& 92.6 & 88.3 & \underline{87.4} & \underline{83.1} & \underline{89.4} \\

\midrule

MPCA
& 90.0 & 90.0 & \textbf{86.5} & \textbf{85.5} & \underline{83.5} 
& 97.0 & \underline{95.5} & \underline{77.5} & \textbf{80.0} & \textbf{92.0} 
& 98.0 & \textbf{88.0} & \textbf{97.0} & \textbf{95.0} & \textbf{97.5} 
& \textbf{96.0} & \textbf{89.0} & \textbf{97.0} & \underline{81.5} & \underline{90.0} 
& 95.3 & \textbf{90.6} & \textbf{89.5} & \textbf{85.5} & \textbf{90.8} \\

\bottomrule
\end{tabular}
}
\label{tab:improved-performance}
\end{table}

\subsubsection{Training Strategies Evaluation}
\label{subsubsec:training-strategies-eval}

Based on the findings in Section~\ref{subsec:main-results}, we provide multiple variants to further improve the multilingual performance. To ensure a fair comparison, we introduce a COCO-VQA dataset provided by StarVLA~\citep{community2026starvla} for VLA cotraining, and augment the dataset with multilingual instructions through translation. The dataset contains 50K image-question-answer triplets, covering a wide range of visual concepts. The multilingual instructions are generated by translating the original English questions into Chinese, French, Russian, and Arabic, using the Cloud Translation API.

We first follow the findings 1 and 4 to choose Qwen3-VL as the base VLM, which has better multilingual performance \citep{bai2025qwen3}, and adopt the GR00T-style action head design for all models. We then fine-tune the models on the multilingual COCO-VQA dataset to improve the multilingual understanding. Two multilingual variants are trained with different fine-tuning strategies: (1) M-FT: multilingual fine-tuning for the base VLM at first, and then training the action head with the LIBERO dataset; (2) M-CT: cotraining the VLA model with both the multilingual COCO-VQA dataset and the LIBERO dataset. We also include two monolingual variants as baselines, denoted as E-FT and E-CT, which are trained with the original English COCO-VQA dataset. The absolute success rate of these variants are provided in Table~\ref{tab:improved-performance}. The original performance of Qwen3-VL-GR00T is also included as a reference, denoted as GR00T.

We analyze the results from three perspectives and provide our findings for each perspective. (1) Compared to the original GR00T, M-FT performs better on multilingual instructions, which further confirms our finding 1 in Section \ref{subsec:main-results} that the multilingual training data in the base VLM can impact the multilingual performance of the resulting VLA models. (2) The average performance of M-FT and M-CT under English and Chinese instructions are generally worse than the original GR00T. It means that the multilingual gains come with a cost of performance drop on English instructions, which is a common phenomenon in multilingual large language models~\citep{qin2025survey, marchisio2024does, dou2023multispider} as well. (3) M-CT performs better than M-FT on multilingual instructions, which suggests that cotraining with both the multilingual COCO-VQA dataset and the LIBERO dataset can help the model better understand and follow multilingual instructions.

\subsubsection{Multilingual Principal Component Alignment}
\label{subsubsec:mpca}

Inspired by the analysis in Section~\ref{subsubsec:representation-shift-analysis} and Section~\ref{subsubsec:training-strategies-eval}, we further propose a simple yet effective method based on M-CT to improve the multilingual performance by aligning the principal components of multilingual instruction embeddings so that the multilingual instruction embeddings can better align with English instruction embeddings.

Specifically, we first sample $n$ groups of samples from the multilingual COCO-VQA dataset, where each group contains the same instruction in different languages. For sample $i$ in language ${j}$, we compute the average pooled embedding $\mathbf{h}_{i, j}$ from the middle layer of the base VLM, and perform principal component analysis (PCA) on these embeddings to obtain the principal components $\mathbf{U}$:
\begin{equation}
    \mathbf{U} = \text{PCA}(\{\mathbf{h}_{i, j}\} | i=1,...,n; j\in\{\text{en}, \text{zh}, \text{fr}, \text{ru}, \text{ar}\})
\end{equation}
We update the principal components $\mathbf{U}$ per $k$ steps during training. We then use the cosine similarity to align the projection of the multilingual instruction embeddings on the principal components with that of English instructions. The loss function is defined as follows:
\begin{equation}
    \mathcal{L} = \mathcal{L}_{\text{VLM}} + \mathcal{L}_{\text{VLA}} + \lambda \sum_{i=1}^{n} \sum_{j=1}^{m} (1 - \cos(\mathbf{U}^T \mathbf{h}_{i, j}, \mathbf{U}^T \mathbf{h}_{i, \text{en}}))
\end{equation}
where $\mathcal{L}_{\text{VLM}}$ and $\mathcal{L}_{\text{VLA}}$ are the original loss functions for training the base VLM and the action head, respectively, $\lambda$ is a hyperparameter that controls the weight of the alignment loss, and $m$ is the number of languages. The insight of this method is to encourage the most significant components of the multilingual instruction embeddings to be aligned with those of English instruction embeddings, while allowing the less significant components to capture other information. We denote the model trained with this method as Multilingual Principal Component Alignment (MPCA). The performance is provided in Table~\ref{tab:improved-performance}, which shows that MPCA can further improve the multilingual performance compared to M-CT. We also provide an ablation study on PCA in the Appendix~\ref{subsec:extended-ablation-study-mpca}, which further supports the effectiveness of MPCA.

\section{Conclusion}
\label{sec:conclusion}

In this work, we present a systematic study of multilingual instruction following in VLA models. Through a comprehensive evaluation and extensive experiments across multiple VLA models and robot benchmarks, we reveal a significant multilingual gap that has been largely overlooked in the field. Our analysis demonstrates that this gap would reflect both instruction understanding and action execution, and multilingual instruction-caused representation shifts may contribute to the multilingual gap. Inspired by these findings, we propose Multilingual Principal Component Alignment (MPCA), a simple yet effective approach that leverages PCA-based representation alignment to improve multilingual performance in VLAs.
 

\section*{Acknowledgements}
We thank the StarVLA team for providing a solid foundation for our evaluation. We are also grateful to the authors of the benchmarks and models used in our experiments for openly sharing their work with the research community.

\clearpage
\newpage
\bibliographystyle{assets/plainnat}
\bibliography{arxiv}

@article{kim2024openvla,
  title={Openvla: An open-source vision-language-action model},
  author={Kim, Moo Jin and Pertsch, Karl and Karamcheti, Siddharth and Xiao, Ted and Balakrishna, Ashwin and Nair, Suraj and Rafailov, Rafael and Foster, Ethan and Lam, Grace and Sanketi, Pannag and others},
  journal={arXiv preprint arXiv:2406.09246},
  year={2024}
}

@article{kim2025fine,
  title={Fine-tuning vision-language-action models: Optimizing speed and success},
  author={Kim, Moo Jin and Finn, Chelsea and Liang, Percy},
  journal={arXiv preprint arXiv:2502.19645},
  year={2025}
}

@article{black2024pi_0,
  title={$\pi_0$: A Vision-Language-Action Flow Model for General Robot Control},
  author={Black, Kevin and Brown, Noah and Driess, Danny and Esmail, Adnan and Equi, Michael and Finn, Chelsea and Fusai, Niccolo and Groom, Lachy and Hausman, Karol and Ichter, Brian and others},
  journal={arXiv preprint arXiv:2410.24164},
  year={2024}
}

@inproceedings{black2025pi_,
  title={$\pi_{0.5}$: a Vision-Language-Action Model with Open-World Generalization},
  author={Black, Kevin and Brown, Noah and Darpinian, James and Dhabalia, Karan and Driess, Danny and Esmail, Adnan and Equi, Michael Robert and Finn, Chelsea and Fusai, Niccolo and Galliker, Manuel Y and others},
  booktitle={9th Annual Conference on Robot Learning},
  year={2025}
}

@article{team2024octo,
  title={Octo: An open-source generalist robot policy},
  author={Team, Octo Model and Ghosh, Dibya and Walke, Homer and Pertsch, Karl and Black, Kevin and Mees, Oier and Dasari, Sudeep and Hejna, Joey and Kreiman, Tobias and Xu, Charles and others},
  journal={arXiv preprint arXiv:2405.12213},
  year={2024}
}

@article{cen2025rynnvla,
  title={Rynnvla-002: A unified vision-language-action and world model},
  author={Cen, Jun and Huang, Siteng and Yuan, Yuqian and Li, Kehan and Yuan, Hangjie and Yu, Chaohui and Jiang, Yuming and Guo, Jiayan and Li, Xin and Luo, Hao and others},
  journal={arXiv preprint arXiv:2511.17502},
  year={2025}
}

@article{wang2024qwen2,
  title={Qwen2-vl: Enhancing vision-language model's perception of the world at any resolution},
  author={Wang, Peng and Bai, Shuai and Tan, Sinan and Wang, Shijie and Fan, Zhihao and Bai, Jinze and Chen, Keqin and Liu, Xuejing and Wang, Jialin and Ge, Wenbin and others},
  journal={arXiv preprint arXiv:2409.12191},
  year={2024}
}

@article{bai2025qwen3,
  title={Qwen3-vl technical report},
  author={Bai, Shuai and Cai, Yuxuan and Chen, Ruizhe and Chen, Keqin and Chen, Xionghui and Cheng, Zesen and Deng, Lianghao and Ding, Wei and Gao, Chang and Ge, Chunjiang and others},
  journal={arXiv preprint arXiv:2511.21631},
  year={2025}
}

@article{hurst2024gpt,
  title={Gpt-4o system card},
  author={Hurst, Aaron and Lerer, Adam and Goucher, Adam P and Perelman, Adam and Ramesh, Aditya and Clark, Aidan and Ostrow, AJ and Welihinda, Akila and Hayes, Alan and Radford, Alec and others},
  journal={arXiv preprint arXiv:2410.21276},
  year={2024}
}

@article{team2024gemini,
  title={Gemini 1.5: Unlocking multimodal understanding across millions of tokens of context},
  author={Team, Gemini and Georgiev, Petko and Lei, Ving Ian and Burnell, Ryan and Bai, Libin and Gulati, Anmol and Tanzer, Garrett and Vincent, Damien and Pan, Zhufeng and Wang, Shibo and others},
  journal={arXiv preprint arXiv:2403.05530},
  year={2024}
}

@article{xu2025seeing,
  title={Seeing to Act, Prompting to Specify: A Bayesian Factorization of Vision Language Action Policy},
  author={Xu, Kechun and Zhu, Zhenjie and Chen, Anzhe and Zhao, Shuqi and Huang, Qing and Yang, Yifei and Lu, Haojian and Xiong, Rong and Tomizuka, Masayoshi and Wang, Yue},
  journal={arXiv preprint arXiv:2512.11218},
  year={2025}
}

@inproceedings{zhang2025vlabench,
  title={Vlabench: A large-scale benchmark for language-conditioned robotics manipulation with long-horizon reasoning tasks},
  author={Zhang, Shiduo and Xu, Zhe and Liu, Peiju and Yu, Xiaopeng and Li, Yuan and Gao, Qinghui and Fei, Zhaoye and Yin, Zhangyue and Wu, Zuxuan and Jiang, Yu-Gang and others},
  booktitle={Proceedings of the IEEE/CVF International Conference on Computer Vision},
  pages={11142--11152},
  year={2025}
}

@article{karnik2024embodied,
  title={Embodied red teaming for auditing robotic foundation models},
  author={Karnik, Sathwik and Hong, Zhang-Wei and Abhangi, Nishant and Lin, Yen-Chen and Wang, Tsun-Hsuan and Dupuy, Christophe and Gupta, Rahul and Agrawal, Pulkit},
  journal={arXiv preprint arXiv:2411.18676},
  year={2024}
}

@article{zhou2025libero,
  title={LIBERO-PRO: Towards Robust and Fair Evaluation of Vision-Language-Action Models Beyond Memorization},
  author={Zhou, Xueyang and Xu, Yangming and Tie, Guiyao and Chen, Yongchao and Zhang, Guowen and Chu, Duanfeng and Zhou, Pan and Sun, Lichao},
  journal={arXiv preprint arXiv:2510.03827},
  year={2025}
}

@article{fei2025libero,
  title={Libero-plus: In-depth robustness analysis of vision-language-action models},
  author={Fei, Senyu and Wang, Siyin and Shi, Junhao and Dai, Zihao and Cai, Jikun and Qian, Pengfang and Ji, Li and He, Xinzhe and Zhang, Shiduo and Fei, Zhaoye and others},
  journal={arXiv preprint arXiv:2510.13626},
  year={2025}
}

@article{liu2023libero,
  title={Libero: Benchmarking knowledge transfer for lifelong robot learning},
  author={Liu, Bo and Zhu, Yifeng and Gao, Chongkai and Feng, Yihao and Liu, Qiang and Zhu, Yuke and Stone, Peter},
  journal={Advances in Neural Information Processing Systems},
  volume={36},
  pages={44776--44791},
  year={2023}
}

@article{li2024evaluating,
  title={Evaluating real-world robot manipulation policies in simulation},
  author={Li, Xuanlin and Hsu, Kyle and Gu, Jiayuan and Pertsch, Karl and Mees, Oier and Walke, Homer Rich and Fu, Chuyuan and Lunawat, Ishikaa and Sieh, Isabel and Kirmani, Sean and others},
  journal={arXiv preprint arXiv:2405.05941},
  year={2024}
}

@article{mees2022calvin,
  title={Calvin: A benchmark for language-conditioned policy learning for long-horizon robot manipulation tasks},
  author={Mees, Oier and Hermann, Lukas and Rosete-Beas, Erick and Burgard, Wolfram},
  journal={IEEE Robotics and Automation Letters},
  volume={7},
  number={3},
  pages={7327--7334},
  year={2022},
  publisher={IEEE}
}

@article{chen2025robotwin,
  title={Robotwin 2.0: A scalable data generator and benchmark with strong domain randomization for robust bimanual robotic manipulation},
  author={Chen, Tianxing and Chen, Zanxin and Chen, Baijun and Cai, Zijian and Liu, Yibin and Li, Zixuan and Liang, Qiwei and Lin, Xianliang and Ge, Yiheng and Gu, Zhenyu and others},
  journal={arXiv preprint arXiv:2506.18088},
  year={2025}
}

@article{nasiriany2024robocasa,
  title={Robocasa: Large-scale simulation of everyday tasks for generalist robots},
  author={Nasiriany, Soroush and Maddukuri, Abhiram and Zhang, Lance and Parikh, Adeet and Lo, Aaron and Joshi, Abhishek and Mandlekar, Ajay and Zhu, Yuke},
  journal={arXiv preprint arXiv:2406.02523},
  year={2024}
}

@inproceedings{li2023behavior,
  title={Behavior-1k: A benchmark for embodied ai with 1,000 everyday activities and realistic simulation},
  author={Li, Chengshu and Zhang, Ruohan and Wong, Josiah and Gokmen, Cem and Srivastava, Sanjana and Mart{\'\i}n-Mart{\'\i}n, Roberto and Wang, Chen and Levine, Gabrael and Lingelbach, Michael and Sun, Jiankai and others},
  booktitle={Conference on Robot Learning},
  pages={80--93},
  year={2023},
  organization={PMLR}
}

@article{brohan2022rt,
  title={Rt-1: Robotics transformer for real-world control at scale},
  author={Brohan, Anthony and Brown, Noah and Carbajal, Justice and Chebotar, Yevgen and Dabis, Joseph and Finn, Chelsea and Gopalakrishnan, Keerthana and Hausman, Karol and Herzog, Alex and Hsu, Jasmine and others},
  journal={arXiv preprint arXiv:2212.06817},
  year={2022}
}

@article{wen2024diffusion,
  title={Diffusion-vla: Generalizable and interpretable robot foundation model via self-generated reasoning},
  author={Wen, Junjie and Zhu, Minjie and Zhu, Yichen and Tang, Zhibin and Li, Jinming and Zhou, Zhongyi and Li, Chengmeng and Liu, Xiaoyu and Peng, Yaxin and Shen, Chaomin and others},
  journal={arXiv preprint arXiv:2412.03293},
  year={2024}
}

@article{bjorck2025gr00t,
  title={Gr00t n1: An open foundation model for generalist humanoid robots},
  author={Bjorck, Johan and Casta{\~n}eda, Fernando and Cherniadev, Nikita and Da, Xingye and Ding, Runyu and Fan, Linxi and Fang, Yu and Fox, Dieter and Hu, Fengyuan and Huang, Spencer and others},
  journal={arXiv preprint arXiv:2503.14734},
  year={2025}
}

@article{bu2025univla,
  title={Univla: Learning to act anywhere with task-centric latent actions},
  author={Bu, Qingwen and Yang, Yanting and Cai, Jisong and Gao, Shenyuan and Ren, Guanghui and Yao, Maoqing and Luo, Ping and Li, Hongyang},
  journal={arXiv preprint arXiv:2505.06111},
  year={2025}
}

@article{cen2025worldvla,
  title={Worldvla: Towards autoregressive action world model},
  author={Cen, Jun and Yu, Chaohui and Yuan, Hangjie and Jiang, Yuming and Huang, Siteng and Guo, Jiayan and Li, Xin and Song, Yibing and Luo, Hao and Wang, Fan and others},
  journal={arXiv preprint arXiv:2506.21539},
  year={2025}
}

@article{lian2026bayesianvla,
  title={BayesianVLA: Bayesian Decomposition of Vision Language Action Models via Latent Action Queries},
  author={Lian, Shijie and Yu, Bin and Lin, Xiaopeng and Yang, Laurence T and Shen, Zhaolong and Wu, Changti and Miao, Yuzhuo and Huang, Cong and Chen, Kai},
  journal={arXiv preprint arXiv:2601.15197},
  year={2026}
}

@article{wang2025vlatest,
  title={Vlatest: Testing and evaluating vision-language-action models for robotic manipulation},
  author={Wang, Zhijie and Zhou, Zhehua and Song, Jiayang and Huang, Yuheng and Shu, Zhan and Ma, Lei},
  journal={Proceedings of the ACM on Software Engineering},
  volume={2},
  number={FSE},
  pages={1615--1638},
  year={2025},
  publisher={ACM New York, NY, USA}
}

@article{lian2026langforce,
  title={LangForce: Bayesian Decomposition of Vision Language Action Models via Latent Action Queries},
  author={Lian, Shijie and Yu, Bin and Lin, Xiaopeng and Yang, Laurence T and Shen, Zhaolong and Wu, Changti and Miao, Yuzhuo and Huang, Cong and Chen, Kai},
  journal={arXiv e-prints},
  pages={arXiv--2601},
  year={2026}
}

@article{bai2025qwen2,
  title={Qwen2. 5-vl technical report, 2025},
  author={Bai, Shuai and Chen, Keqin and Liu, Xuejing and Wang, Jialin and Ge, Wenbin and Song, Sibo and Dang, Kai and Wang, Peng and Wang, Shijie and Tang, Jun and others},
  journal={URL https://arxiv. org/abs/2502.13923},
  volume={6},
  pages={13--23},
  year={2025}
}

@article{intelligence2025pi_,
  title={$\pi_{0.5}$: a Vision-Language-Action Model with Open-World Generalization},
  author={Intelligence, Physical and Black, Kevin and Brown, Noah and Darpinian, James and Dhabalia, Karan and Driess, Danny and Esmail, Adnan and Equi, Michael and Finn, Chelsea and Fusai, Niccolo and others},
  journal={arXiv preprint arXiv:2504.16054},
  year={2025}
}

@article{yang2026abot,
  title={Abot-m0: Vla foundation model for robotic manipulation with action manifold learning},
  author={Yang, Yandan and Zeng, Shuang and Lin, Tong and Chang, Xinyuan and Qi, Dekang and Xiao, Junjin and Liu, Haoyun and Chen, Ronghan and Chen, Yuzhi and Huo, Dongjie and others},
  journal={arXiv preprint arXiv:2602.11236},
  year={2026}
}

@article{community2026starvla,
  title={StarVLA: A Lego-like Codebase for Vision-Language-Action Model Developing},
  author={Community, StarVLA},
  journal={arXiv preprint arXiv:2604.05014},
  year={2026}
}

@article{kim2026cosmos,
  title={Cosmos policy: Fine-tuning video models for visuomotor control and planning},
  author={Kim, Moo Jin and Gao, Yihuai and Lin, Tsung-Yi and Lin, Yen-Chen and Ge, Yunhao and Lam, Grace and Liang, Percy and Song, Shuran and Liu, Ming-Yu and Finn, Chelsea and others},
  journal={arXiv preprint arXiv:2601.16163},
  year={2026}
}

@article{torne2026mem,
  title={Mem: Multi-scale embodied memory for vision language action models},
  author={Torne, Marcel and Pertsch, Karl and Walke, Homer and Vedder, Kyle and Nair, Suraj and Ichter, Brian and Ren, Allen Z and Wang, Haohuan and Tang, Jiaming and Stachowicz, Kyle and others},
  journal={arXiv preprint arXiv:2603.03596},
  year={2026}
}

@inproceedings{karamcheti2024prismatic,
  title={Prismatic vlms: Investigating the design space of visually-conditioned language models},
  author={Karamcheti, Siddharth and Nair, Suraj and Balakrishna, Ashwin and Liang, Percy and Kollar, Thomas and Sadigh, Dorsa},
  booktitle={Forty-first International Conference on Machine Learning},
  year={2024}
}

@article{thellmann2024towards,
  title={Towards multilingual llm evaluation for european languages},
  author={Thellmann, Klaudia and Stadler, Bernhard and Fromm, Michael and Buschhoff, Jasper Schulze and Jude, Alex and Barth, Fabio and Leveling, Johannes and Flores-Herr, Nicolas and K{\"o}hler, Joachim and J{\"a}kel, Ren{\'e} and others},
  journal={arXiv preprint arXiv:2410.08928},
  year={2024}
}

@inproceedings{yong2025state,
  title={The state of multilingual llm safety research: From measuring the language gap to mitigating it},
  author={Yong, Zheng-Xin and Ermis, Beyza and Fadaee, Marzieh and Bach, Stephen and Kreutzer, Julia},
  booktitle={Proceedings of the 2025 Conference on Empirical Methods in Natural Language Processing},
  pages={15856--15871},
  year={2025}
}

@article{luo2026lost,
  title={Lost in Execution: On the Multilingual Robustness of Tool Calling in Large Language Models},
  author={Luo, Zheng and Kutralingam, T Pranav and Okoani, Ogochukwu N and Xu, Wanpeng and Wei, Hua and Hu, Xiyang},
  journal={arXiv preprint arXiv:2601.05366},
  year={2026}
}

@article{qin2025survey,
  title={A survey of multilingual large language models},
  author={Qin, Libo and Chen, Qiguang and Zhou, Yuhang and Chen, Zhi and Li, Yinghui and Liao, Lizi and Li, Min and Che, Wanxiang and Yu, Philip S},
  journal={Patterns},
  volume={6},
  number={1},
  year={2025},
  publisher={Elsevier}
}

@inproceedings{marchisio2024does,
  title={How does quantization affect multilingual LLMs?},
  author={Marchisio, Kelly and Dash, Saurabh and Chen, Hongyu and Aumiller, Dennis and {\"U}st{\"u}n, Ahmet and Hooker, Sara and Ruder, Sebastian},
  booktitle={Findings of the Association for Computational Linguistics: EMNLP 2024},
  pages={15928--15947},
  year={2024}
}

@inproceedings{dou2023multispider,
  title={MultiSpider: towards benchmarking multilingual text-to-SQL semantic parsing},
  author={Dou, Longxu and Gao, Yan and Pan, Mingyang and Wang, Dingzirui and Che, Wanxiang and Zhan, Dechen and Lou, Jian-Guang},
  booktitle={Proceedings of the AAAI Conference on Artificial Intelligence},
  volume={37},
  number={11},
  pages={12745--12753},
  year={2023}
}

@article{steiner2024paligemma,
  title={Paligemma 2: A family of versatile vlms for transfer},
  author={Steiner, Andreas and Pinto, Andr{\'e} Susano and Tschannen, Michael and Keysers, Daniel and Wang, Xiao and Bitton, Yonatan and Gritsenko, Alexey and Minderer, Matthias and Sherbondy, Anthony and Long, Shangbang and others},
  journal={arXiv preprint arXiv:2412.03555},
  year={2024}
}

\clearpage
\newpage
\beginappendix
\section{Limitations}
\label{sec:limitations}

While our work provides valuable insights into the multilingual capabilities of VLA models and proposes an effective approach to improve multilingual performance, there are several limitations that should be acknowledged. First, our evaluation is limited to a specific set of VLA models and robot benchmarks, which may not fully capture the diversity of real-world scenarios and applications. Future work could explore a wider range of models and benchmarks to further validate our findings. Second, while MPCA shows promising results in improving multilingual performance, it is a relatively simple approach that may not fully address all the challenges associated with multilingual instruction following. More sophisticated methods that consider the nuances of different languages and their interactions with visual and action representations could be explored in future research. Third, our analysis primarily focuses on the representation shift caused by multilingual instructions, but other factors, such as data quality, may also contribute to the multilingual gap. A more comprehensive analysis that considers these factors could provide deeper insights into the underlying causes of the multilingual gap and inform more effective solutions.
\section{Broader Impacts}
\label{sec:broader-impacts}

Our work has several broader impacts that are important to consider. First, by systematically studying multilingual instruction following in VLA models, we highlight the importance of multilingual capabilities in embodied AI systems. This can encourage researchers and practitioners to prioritize multilingual performance when developing and deploying VLA models, leading to more inclusive and accessible technologies that can serve a wider range of users across different languages and cultures. Second, our findings regarding the multilingual gap and its underlying causes can inform the design of future VLA models and training strategies, potentially leading to more robust and effective multilingual instruction following capabilities. This can have positive impacts on the usability and effectiveness of VLA systems in real-world applications, such as assistive robotics, where users may interact with robots in their native languages. Third, our proposed MPCA approach offers a practical solution for improving multilingual performance in VLAs, which can be adopted by researchers and practitioners to enhance the multilingual robustness of their models. This can contribute to the development of more versatile and adaptable VLA systems that can better serve diverse user needs. However, it is also important to acknowledge potential risks associated with multilingual VLA systems, such as the possibility of biased performance across different languages or the misuse of multilingual capabilities in harmful ways. Future research should continue to explore these risks and develop strategies to mitigate them, ensuring that the benefits of multilingual VLA systems are realized while minimizing potential harms. Overall, our work contributes to advancing the field of embodied AI towards more inclusive and multilingual-aware systems, with the potential to positively impact a wide range of applications and users.
\section{Dataset Construction Details}
\label{sec:dataset-construction}

In this section, we provide detailed information about the construction of our multilingual instruction-following dataset. 

\textbf{Multilingual Instruction Construction.}
We utilize the v2 version of Cloud Translation API~\footnote{https://translation.googleapis.com/language/translate/v2} to translate the original English instructions into the target languages.

\textbf{Code-switching Instruction Construction.}
To construct code-switching instructions, we leverage the multilingual capabilities of LLMs to identify key verbs and nouns in both the original English instructions and their translated versions. We then substitute the corresponding key phrases in the translated instructions with those from the original English instructions, resulting in code-switched instructions that contain mixed-language expressions. Another LLM is used to evaluate the semantic consistency between the original English instructions and the code-switching instructions, ensuring that the generated instructions maintain their intended meaning. The specific prompt templates used for code-switching instruction generation and evaluation are provided. We use gpt-5.2-20251211 for both generation and evaluation, with temperature set to 0.2.

\begin{tcolorbox}[
  colframe=citecolor!50!black,
  colback=citecolor!10!white,
  boxrule=1.5pt,
  enhanced,
  arc=2pt,
  title=Code-Switching Prompt Template,
  fontupper=\normalsize,
  fonttitle=\bfseries\normalsize,
  left=3mm, right=3mm, top=2mm, bottom=2mm,
  before skip=5pt, after skip=5pt
]

You are a code-mixing assistant for multilingual instruction augmentation.

Given English tokens [EN] and target language tokens [LT], your task is to generate a code-mixed variant of [LT] by replacing 1–3 nouns or verbs with tokens from [EN].

Follow these rules strictly:

\textbf{Procedure}
\begin{itemize}
  \item Identify tokens in [LT] with POS $\in \{\text{NOUN, VERB}\}$.
  \item Uniformly sample $k \in \{1,2,3\}$ tokens.
  \item Replace each sampled token with a token from [EN].
  \item Keep all other tokens unchanged.
\end{itemize}

\textbf{Constraints}
\begin{itemize}
  \item Do not introduce tokens outside [EN].
  \item Do not modify tokens outside [LT].
  \item Preserve word order and fluency.
\end{itemize}

\textbf{Output Format}
\begin{itemize}
  \item The output must start with \#\#\#.
  \item Format strictly as: \#\#\# \texttt{<Modified target language instruction>}
  \item Do not include explanations or extra text.
\end{itemize}

\label{temp:code-switching-prompt}
\end{tcolorbox}

\begin{tcolorbox}[
  colframe=citecolor!50!black,
  colback=citecolor!10!white,
  boxrule=1.5pt,
  enhanced,
  arc=2pt,
  title=Code-Switching Evaluation Template,
  fontupper=\normalsize,
  fonttitle=\bfseries\normalsize,
  left=3mm, right=3mm, top=2mm, bottom=2mm,
  before skip=5pt, after skip=5pt,
]

You are a strict evaluator of code-mixed instructions for a multilingual VLA dataset.

Given English tokens [EN] and target language tokens [LT], your task is to evaluate whether a code-mixed instruction [CANDIDATE] satisfies the following criteria.

Rules the [CANDIDATE] MUST satisfy:
\begin{itemize}
  \item The candidate equals [LT] except 1–3 tokens are replaced with English words from [EN].
  \item Replaced tokens are NOUN or VERB in [LT].
  \item No tokens are introduced that are absent from both [LT] and [EN] (e.g. extra English articles or particles like 'the', 'a').
  \item Word order is preserved.
  \item Semantic meaning is identical to [EN].
\end{itemize}

Think briefly (1–3 short sentences), then output the verdict on the LAST line strictly as `\#\#\# YES` or `\#\#\# NO`. Nothing after the verdict line.

\label{temp:code-switching-evaluation}
\end{tcolorbox}
\section{Experiment Details}
\label{sec:experiment-details}

\subsection{Model Details}
\label{subsec:model-details}

In this section, we provide detailed information about the models evaluated in our experiments. Models we include can be categorized into two groups: (1) models trained primarily on English instructions, including OpenVLA-OFT~\citep{kim2025fine}, $\pi_{0.5}$~\citep{intelligence2025pi_}, and Cosmos Policy~\citep{kim2026cosmos}; (2) Qwen-VL-based VLAs with different action head designs~\citep{community2026starvla}, which are evaluated to understand the impact of multilingual VLM backbones and architectural choices on multilingual performance. The specific details of each model are provided below:
\begin{itemize}
    \item \textbf{OpenVLA-OFT}~\citep{kim2025fine} is a VLA model that utilizes LLaMA-2-7B as the backbone and is trained on a large-scale dataset. It employs an Optimized Finetuning (OFT) approach to enhance the capabilities of the model. We use the official checkpoint finetuned on LIBERO for evaluation, which is available at \url{https://huggingface.co/moojink/openvla-7b-oft-finetuned-libero-spatial-object-goal-10}.
    \item \textbf{$\pi_{0.5}$}~\citep{intelligence2025pi_} is a VLA model that leverages a diffusion-based continuous action expert to generate actions. We use the official checkpoint finetuned on LIBERO for evaluation, which is available at \url{https://storage.googleapis.com/openpi-assets/checkpoints/pi05_libero}.
    \item \textbf{ABot-M0}~\citep{yang2026abot} is a VLA model that incorporates a manifold-based action head to enhance the model's ability to generate actions. We use the official checkpoint finetuned on LIBERO for evaluation, which is available at \url{https://huggingface.co/acvlab/ABot-M0-LIBERO}.
    \item \textbf{Cosmos Policy}~\citep{kim2026cosmos} is a world-model-based policy that utilizes a video-based world model to predict future states and generate actions. We use the official checkpoint finetuned on LIBERO for evaluation, which is available at \url{https://huggingface.co/nvidia/Cosmos-Policy-LIBERO-Predict2-2B}.
    \item \textbf{Qwen-VL-based VLAs}~\citep{community2026starvla} are a series of VLA models that utilize Qwen-VL as the backbone and incorporate different action head designs. We evaluate multiple variants of these models to understand the impact of multilingual VLM backbones and architectural choices on multilingual performance. We use the official checkpoints finetuned on LIBERO or SimplerEnv for evaluation, which are available at \url{https://huggingface.co/StarVLA}.
\end{itemize}

\subsection{Training Details}

In this section, we provide training details of MPCA and other multilingual variants. For MPCA, E-CT, and M-CT, we use a batch size of 2 per GPU and train for 50K steps on 8 A800 GPUs or higher-end GPUs. The learning rate is set to 2.5e-5, with a cosine learning rate schedule. We use the AdamW optimizer with no weight decay. For MPCA, we set the number of principal components to 128 and the alignment weight to 0.01. The principal components will be updated every 1000 steps. For E-CT and M-CT, we use the same training data and hyperparameters as MPCA, which consists of multilingual instruction variants generated from the original English instructions in the respective benchmarks.

For E-FT and M-FT, we keep the same training settings as MPCA in the VLA training stage. But it should be noted that we adopt the batch size of 8 per GPU for E-FT and M-FT in the VLM fine-tuning stage, which is different from the batch size used in MPCA. The training steps for E-FT and M-FT are set to 30K steps in the VLM fine-tuning stage and 50K steps in the VLA training stage.
\section{Extended Experimental Results and Analysis}
\label{sec:extended-experimental-results}

\subsection{Extended Main Results}
\label{subsec:extended-main-results}

We further provide the absolute performance of all models in a multilingual instruction setting in Table~\ref{tab:multilingual-results-absolute} and Table~\ref{tab:multilingual-results-absolute-simpler-env}. We also provide the absolute code-switching performance in Table~\ref{tab:code-switching-absolute-results} and Table~\ref{tab:code-switching-absolute-results-simpler-env}.

\begin{table}[H]
\caption{Multilingual performance on LIBERO across four suites. Absolute performance is shown. $\varnothing$ denotes evaluated without any instructions. Percentage sign is omitted for better readability.}
\centering
\small
\setlength{\tabcolsep}{3.5pt}
\renewcommand{\arraystretch}{1.15}
\resizebox{\textwidth}{!}{
\begin{tabular}{l|c|cccccc|cccccc|cccccc|cccccc}
\toprule

\multicolumn{2}{c|}{\multirow{2}{*}{Models}}
& \multicolumn{6}{c|}{Long}
& \multicolumn{6}{c|}{Goal}
& \multicolumn{6}{c|}{Object}
& \multicolumn{6}{c}{Spatial} \\
\cmidrule(lr){3-8} \cmidrule(lr){9-14} \cmidrule(lr){15-20} \cmidrule(lr){21-26}

\multicolumn{2}{c|}{}
& en & zh & fr & ru & ar & $\varnothing$
& en & zh & fr & ru & ar & $\varnothing$
& en & zh & fr & ru & ar & $\varnothing$
& en & zh & fr & ru & ar & $\varnothing$ \\

\midrule

\multicolumn{2}{c|}{OpenVLA-OFT}
& 94.5 & 89.0 & 89.5 & 80.5 & 79.0 & 76.5 
& 97.5 & 64.0 & 9.5 & 8.0 & 10.5 & 10.0 
& 100.0 & 95.5 & 81.0 & 88.5 & 91.5 & 91.5 
& 93.5 & 90.0 & 84.5 & 86.0 & 78.0 & 80.0 \\

\multicolumn{2}{c|}{$\pi_{0.5}$}
& 93.0 & 73.0 & 76.0 & 69.5 & 72.0 & 70.0 & 94.0 & 16.0 & 9.5 & 5.5 & 8.5 & 9.0 & 99.0 & 66.5 & 71.0 & 67.5 & 65.5 & 64.5 & 97.5 & 69.5 & 75.0 & 74.0 & 72.0 & 66.0 \\

\multicolumn{2}{c|}{ABot-M0}
& 94.5 & 89.0 & 89.5 & 80.5 & 79.0 & 76.5 & 97.5 & 64.0 & 9.5 & 8.0 & 10.5 & 10.0 & 100.0 & 95.5 & 81.0 & 88.5 & 91.5 & 91.5 & 93.5 & 90.0 & 84.5 & 86.0 & 78.0 & 80.0 \\

\multicolumn{2}{c|}{Cosmos Policy}
& 98.0 & 78.5 & 94.5 & 83.5 & 77.5 & 76.0 & 96.5 & 7.5 & 50.0 & 15.0 & 9.0 & 9.0 & 99.5 & 61.5 & 97.5 & 71.5 & 61.5 & 58.5 & 96.5 & 57.5 & 80.5 & 65.5 & 56.5 & 55.5 \\

\midrule
{\footnotesize\multirow{4}{*}{\rotatebox{90}{\textbf{Qwen2.5-VL}}}}
& OFT
& 89.5 & 82.5 & 67.5 & 63.5 & 54.0 & 56.6 & 97.5 & 79.5 & 26.0 & 26.0 & 7.5 & 5.5 & 98.5 & 79.5 & 90.0 & 64.0 & 56.5 & 75.1 & 91.5 & 64.5 & 65.0 & 42.5 & 22.5 & 66.5 \\

& FAST
& 88.5 & 66.0 & 37.5 & 33.5 & 36.5 & 36.0 & 91.0 & 71.0 & 7.5 & 11.0 & 7.5 & 8.0 & 96.5 & 66.5 & 50.5 & 49.5 & 48.0 & 50.0 & 86.5 & 55.5 & 4.5 & 5.0 & 10.0 & 2.0 \\

& GR00T
& 93.5 & 96.5 & 79.0 & 72.5 & 67.0 & 65.0 & 97.0 & 79.5 & 27.5 & 32.5 & 16.5 & 4.0 & 99.5 & 89.5 & 91.0 & 77.5 & 63.5 & 77.5 & 95.5 & 82.0 & 84.0 & 73.0 & 72.0 & 62.5 \\

& $\pi$
& 89.0 & 72.5 & 66.5 & 58.0 & 62.5 & 60.0 & 94.5 & 37.0 & 18.0 & 15.5 & 11.5 & 11.0 & 99.0 & 77.0 & 72.0 & 71.5 & 63.5 & 60.0 & 88.0 & 53.5 & 61.0 & 49.0 & 51.0 & 52.5 \\

\midrule
{\footnotesize\multirow{4}{*}{\rotatebox{90}{\textbf{Qwen3-VL}}}}
& OFT
& 91.0 & 94.0 & 68.0 & 58.0 & 58.0 & 45.0 & 99.0 & 96.0 & 21.0 & 7.5 & 20.0 & 8.5 & 100.0 & 86.5 & 82.0 & 85.5 & 75.5 & 42.5 & 92.5 & 77.0 & 55.0 & 36.5 & 40.5 & 34.0 \\

& FAST
& 83.0 & 57.5 & 17.5 & 13.0 & 14.0 & 19.5 & 87.5 & 55.5 & 7.5 & 5.0 & 9.0 & 6.5 & 98.0 & 71.5 & 35.0 & 29.5 & 33.5 & 37.0 & 89.0 & 54.5 & 14.5 & 4.5 & 6.5 & 12.5 \\

& GR00T
& 92.5 & 93.5 & 73.0 & 69.5 & 66.5 & 58.5 & 97.5 & 95.5 & 10.5 & 13.5 & 22.5 & 5.0 & 97.0 & 88.0 & 83.0 & 63.0 & 72.0 & 61.0 & 94.0 & 79.5 & 82.5 & 80.0 & 76.0 & 64.5 \\

& $\pi$
& 96.5 & 90.5 & 74.0 & 78.0 & 70.0 & 60.0 & 97.0 & 86.5 & 7.0 & 10.0 & 15.5 & 6.0 & 99.5 & 89.5 & 66.0 & 52.0 & 57.0 & 57.5 & 93.0 & 71.0 & 75.5 & 67.5 & 56.5 & 41.0 \\

\bottomrule
\end{tabular}
}
\label{tab:multilingual-results-absolute}
\end{table}
\begin{table}[H]
\caption{
    Multilingual performance on SimperEnv across four tasks. Absolute performance is shown. $\varnothing$ denotes evaluated without any instructions. Percentage sign is omitted for better readability.
}
\centering
\small
\setlength{\tabcolsep}{3.5pt}
\renewcommand{\arraystretch}{1.15}
\resizebox{\textwidth}{!}{
\begin{tabular}{l|c|cccccc|cccccc|cccccc|cccccc}
\toprule

\multicolumn{2}{c|}{\multirow{2}{*}{Models}}
& \multicolumn{6}{c|}{\makecell{Put Spoon \\ on Towel}}
& \multicolumn{6}{c|}{\makecell{Put Carrot \\ on Plate}}
& \multicolumn{6}{c|}{\makecell{Stack Green Block \\ on Yellow Block}}
& \multicolumn{6}{c}{\makecell{Put Eggplant \\ in Yellow Basket}} \\
\cmidrule(lr){3-8} \cmidrule(lr){9-14} \cmidrule(lr){15-20} \cmidrule(lr){21-26}

\multicolumn{2}{c|}{}
& en & zh & fr & ru & ar & $\varnothing$
& en & zh & fr & ru & ar & $\varnothing$
& en & zh & fr & ru & ar & $\varnothing$
& en & zh & fr & ru & ar & $\varnothing$ \\

\midrule
{\footnotesize\multirow{4}{*}{\rotatebox{90}{\textbf{Qwen2.5-VL}}}}
& OFT
& 37.5 & 4.2 & 12.5 & 12.5 & 16.7 & 0.0 & 33.3 & 12.5 & 16.7 & 16.7 & 12.5 & 0.0 & 8.3 & 0.0 & 0.0 & 0.0 & 0.0 & 0.0 & 87.5 & 29.2 & 37.5 & 37.5 & 29.2 & 0.0 \\

& FAST
& 79.2 & 50.0 & 66.7 & 41.7 & 50.0 & 0.0 & 50.0 & 45.8 & 37.5 & 33.3 & 0.0 & 0.0 & 37.5 & 25.0 & 25.0 & 8.3 & 8.3 & 0.0 & 83.3 & 45.8 & 83.3 & 29.2 & 95.8 & 0.0 \\

& GR00T
& 84.4 & 84.4 & 83.3 & 11.5 & 8.3 & 0.0 & 55.2 & 54.2 & 41.7 & 26.0 & 1.0 & 4.2 & 41.7 & 33.3 & 18.1 & 4.2 & 0.0 & 0.0 & 66.7 & 56.3 & 55.2 & 17.7 & 75.0 & 0.0 \\

& $\pi$
& 79.2 & 91.7 & 83.3 & 16.7 & 0.0 & 0.0 & 79.2 & 62.5 & 45.8 & 54.2 & 20.8 & 0.0 & 29.2 & 16.7 & 12.5 & 16.7 & 0.0 & 0.0 & 62.5 & 83.3 & 70.8 & 87.5 & 12.5 & 0.0 \\

\bottomrule
\end{tabular}
}
\label{tab:multilingual-results-absolute-simpler-env}
\end{table}
\begin{table}[H]
\caption{Code-switching performance on LIBERO across four suites. Absolute performance is shown. Percentage sign is omitted for better readability.}
\centering
\small
\setlength{\tabcolsep}{3.5pt}
\renewcommand{\arraystretch}{1.15}
\resizebox{\textwidth}{!}{
\begin{tabular}{l|c|cccc|cccc|cccc|cccc}
\toprule

\multicolumn{2}{c|}{\multirow{2}{*}{Models}}
& \multicolumn{4}{c|}{Long}
& \multicolumn{4}{c|}{Goal}
& \multicolumn{4}{c|}{Object}
& \multicolumn{4}{c}{Spatial} \\
\cmidrule(lr){3-6} \cmidrule(lr){7-10} \cmidrule(lr){11-14} \cmidrule(lr){15-18}

\multicolumn{2}{c|}{}
& zh & fr & ru & ar
& zh & fr & ru & ar
& zh & fr & ru & ar
& zh & fr & ru & ar \\

\midrule

\multicolumn{2}{c|}{OpenVLA-OFT}
& 79.5 & 83.5 & 84.0 & 90.0
& 26.0 & 54.0 & 45.5 & 48.5
& 97.0 & 99.5 & 98.5 & 98.0
& 81.5 & 91.0 & 87.5 & 80.5 \\

\multicolumn{2}{c|}{$\pi_{0.5}$}
& 89.0 & 90.5 & 84.5 & 81.0
& 77.0 & 66.0 & 62.0 & 60.5
& 86.0 & 97.5 & 92.5 & 80.5
& 76.5 & 81.0 & 77.5 & 74.0 \\

\multicolumn{2}{c|}{ABot-M0}
& 94.0 & 93.0 & 81.5 & 82.0
& 75.5 & 70.5 & 60.0 & 57.5
& 93.5 & 98.5 & 99.0 & 94.0
& 94.5 & 83.0 & 87.5 & 87.0 \\

\multicolumn{2}{c|}{Cosmos Policy}
& 82.50 & 91.50 & 95.00 & 87.00
& 46.50 & 75.50 & 60.50 & 35.50
& 74.50 & 97.00 & 94.50 & 79.00
& 56.50 & 87.50 & 75.00 & 66.50 \\

\midrule
{\footnotesize\multirow{4}{*}{\rotatebox{90}{\textbf{Qwen2.5-VL}}}}
& OFT
& 89.0 & 74.5 & 68.0 & 69.0
& 86.0 & 77.0 & 65.5 & 39.5
& 87.0 & 95.5 & 97.0 & 79.5
& 63.5 & 69.0 & 58.0 & 40.0 \\

& FAST
& 68.5 & 41.0 & 42.0 & 45.5
& 70.5 & 28.0 & 39.5 & 38.0
& 66.5 & 53.5 & 59.0 & 53.0
& 56.0 & 12.5 & 56.0 & 15.5 \\

& GR00T
& 87.0 & 84.5 & 78.5 & 79.5
& 80.0 & 71.0 & 70.0 & 61.5
& 90.5 & 95.0 & 87.5 & 80.5
& 77.0 & 82.0 & 75.0 & 75.5 \\

& $\pi$
& 74.0 & 74.5 & 64.5 & 65.0
& 52.0 & 39.5 & 45.0 & 39.0
& 85.5 & 85.5 & 89.5 & 76.0
& 52.5 & 61.0 & 52.0 & 50.5 \\

\midrule
{\footnotesize\multirow{4}{*}{\rotatebox{90}{\textbf{Qwen3-VL}}}}
& OFT
& 92.0 & 72.5 & 84.0 & 74.5
& 96.0 & 79.5 & 65.0 & 68.0
& 89.5 & 99.5 & 95.0 & 89.5
& 78.0 & 62.0 & 49.5 & 40.5 \\

& FAST
& 57.0 & 19.5 & 23.5 & 23.0
& 64.5 & 32.5 & 35.5 & 41.0
& 61.0 & 54.5 & 53.5 & 43.0
& 52.0 & 27.0 & 10.0 & 12.0 \\

& GR00T
& 94.0 & 78.0 & 81.0 & 72.0
& 98.5 & 67.0 & 62.5 & 67.5
& 88.5 & 96.5 & 96.5 & 89.0
& 79.5 & 86.0 & 78.0 & 81.5 \\

& $\pi$
& 92.5 & 85.5 & 82.0 & 82.0
& 96.0 & 59.0 & 61.5 & 52.0
& 88.5 & 85.5 & 93.0 & 71.0
& 72.5 & 84.0 & 72.5 & 62.0 \\

\bottomrule
\end{tabular}
}
\label{tab:code-switching-absolute-results}
\end{table}
\begin{table}[H]
\caption{
    Code-switching performance on SimperEnv across four tasks. Absolute performance is shown. Percentage sign is omitted for better readability.
}
\centering
\small
\setlength{\tabcolsep}{3.5pt}
\renewcommand{\arraystretch}{1.15}
\resizebox{\textwidth}{!}{
\begin{tabular}{l|c|cccc|cccc|cccc|cccc}
\toprule

\multicolumn{2}{c|}{\multirow{2}{*}{Models}}
& \multicolumn{4}{c|}{\makecell{Put Spoon \\ on Towel}}
& \multicolumn{4}{c|}{\makecell{Put Carrot \\ on Plate}}
& \multicolumn{4}{c|}{\makecell{Stack Green Block \\ on Yellow Block}}
& \multicolumn{4}{c}{\makecell{Put Eggplant \\ in Yellow Basket}} \\
\cmidrule(lr){3-6} \cmidrule(lr){7-10} \cmidrule(lr){11-14} \cmidrule(lr){15-18}

\multicolumn{2}{c|}{}
& zh & fr & ru & ar
& zh & fr & ru & ar
& zh & fr & ru & ar
& zh & fr & ru & ar \\

\midrule
{\footnotesize\multirow{4}{*}{\rotatebox{90}{\textbf{Qwen2.5-VL}}}}
& OFT
& 4.2 & 12.5 & 12.5 & 12.5
& 12.5 & 20.8 & 8.3 & 12.5
& 4.2 & 0.0 & 4.2 & 0.0
& 25.0 & 45.8 & 16.7 & 4.2 \\

& FAST
& 87.5 & 62.5 & 83.3 & 20.8
& 37.5 & 45.8 & 37.5 & 37.5
& 16.7 & 20.8 & 20.8 & 0.0
& 95.8 & 66.7 & 95.8 & 87.5 \\

& GR00T
& 75.0 & 70.8 & 87.5 & 33.3
& 54.2 & 37.5 & 41.7 & 37.5
& 16.7 & 4.2 & 8.3 & 0.0
& 70.8 & 62.5 & 79.2 & 62.5 \\

& $\pi$
& 83.3 & 85.4 & 87.5 & 33.3
& 66.7 & 45.8 & 62.5 & 50.0
& 12.5 & 4.2 & 25.0 & 4.2
& 66.7 & 62.5 & 75.0 & 62.5 \\

\bottomrule
\end{tabular}
}
\label{tab:code-switching-absolute-results-simpler-env}
\end{table}

\subsection{Extended Model Behavior Analysis}
\label{subsec:extended-model-behavior-analysis}

\begin{figure}
    \centering
    \subfloat[Behavior of Qwen2.5-FAST with the instruction in French: "pick up the BBQ sauce and place it in the basket"]{
        \includegraphics[width=0.99\linewidth]{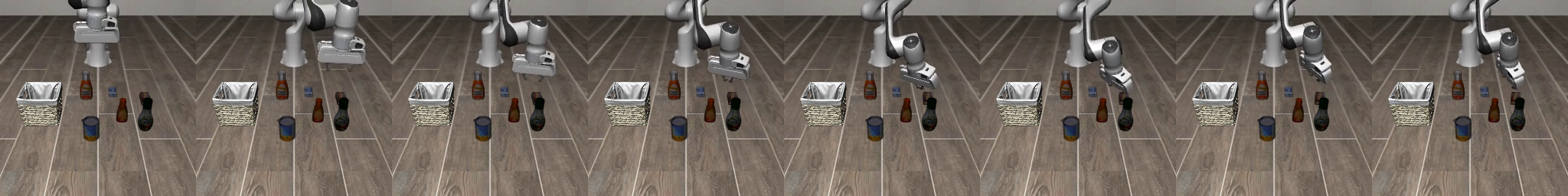}
    }
    \\
    \subfloat[Behavior of OpenVLA-OFT with the instruction in Chinese: "open the middle drawer of the cabinet"]{
        \includegraphics[width=0.99\linewidth]{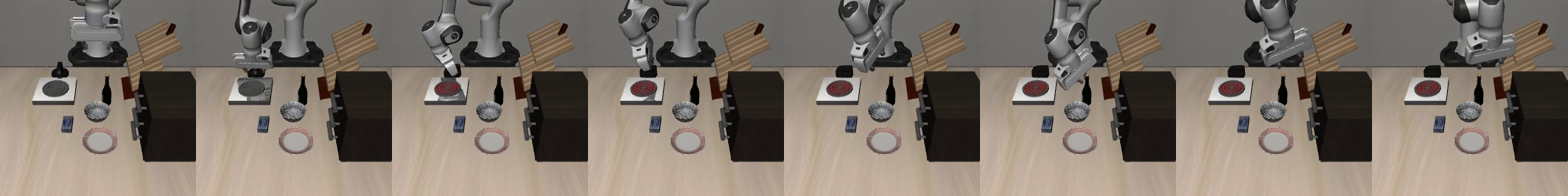}
    }
    \\
    \subfloat[Behavior of ABot-M0 with the instruction in Chinese: "open the top drawer and put the bowl inside"]{
        \includegraphics[width=0.99\linewidth]{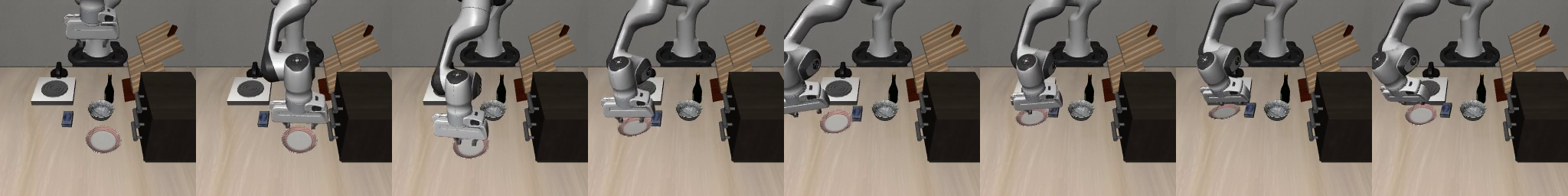}
    }
    \\
    \subfloat[Behavior of Qwen3-GR00T with the instruction in Russian: "put both moka pots on the stove"]{
        \includegraphics[width=0.99\linewidth]{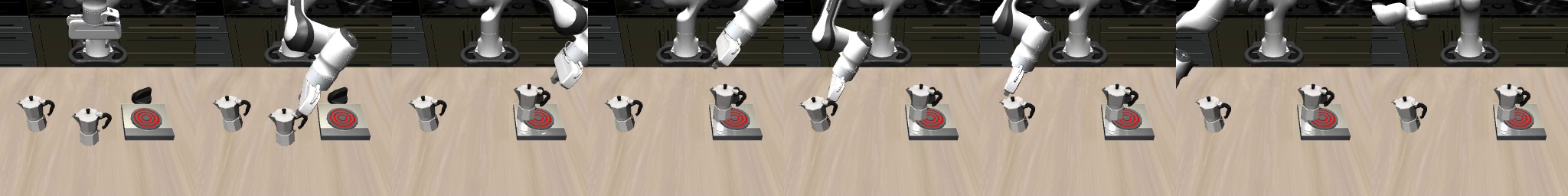}
    }
    \\
    \subfloat[Behavior of Qwen2.5-VL-GR00T with the instruction in Arabic: "put the white mug on the left plate and put the yellow and white mug on the right plate"]{
        \includegraphics[width=0.99\linewidth]{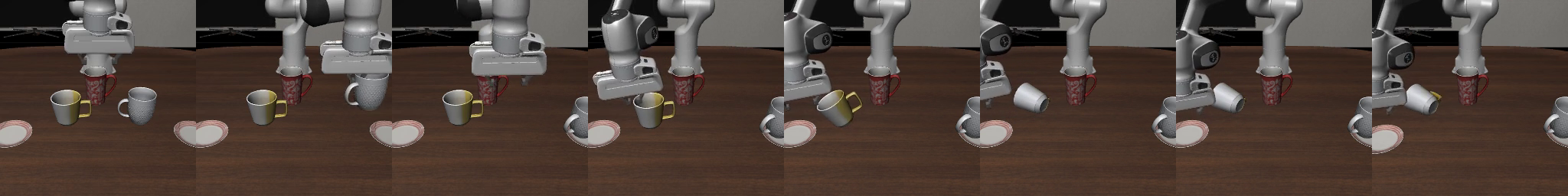}
    }
    \\
    \subfloat[Behavior of Qwen2.5-VL-$\pi$ with the instruction in Arabic: "stack green block on yellow block"]{
        \includegraphics[width=0.99\linewidth]{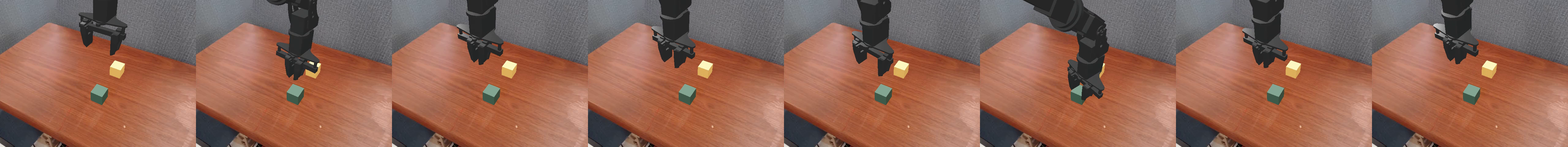}
    }
    \\
    \subfloat[Behavior of Qwen2.5-VL-FAST with the instruction in Russian: "put the spoon on the towel"]{
        \includegraphics[width=0.99\linewidth]{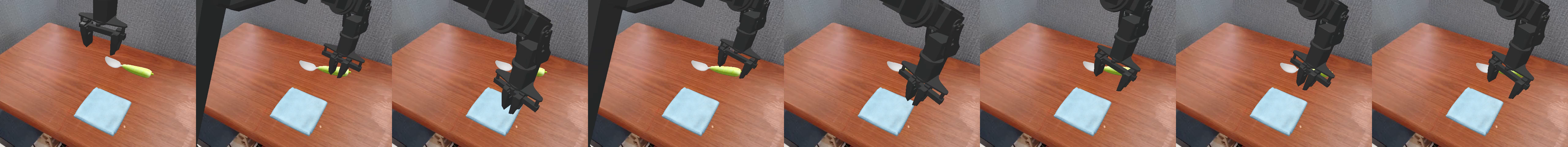}
    }
    \caption{
        Behavioral analysis of the failure cases under multilingual instructions. Case (a)-(e) shows failure cases on LIBERO. Case (f) and (g) shows failure cases on SimperEnv. Case (a)-(c) shows failure cases where the model fails to understand the instruction and thus performs wrong actions. Case (d)-(g) shows failure cases where the model recognizes the instruction but fails to execute the correct action.
    }
    \label{fig:extended-failure-cases}
\end{figure}

We provide additional behavioral analysis of the failure cases under multilingual instructions in Figure~\ref{fig:extended-failure-cases}. The observations confirm our previous analysis in Section~\ref{subsubsec:model-behavior-analysis} that the performance drop in a multilingual instruction setting is not only due to the language understanding capability of the model, but also due to the execution capability of the model under multilingual instructions. In case (d)-(g), models can always recognize the objects or the spatial relationships in the instruction, but fail to execute the correct actions. This indicates that even if the model can understand the instruction, it may still struggle to execute the correct actions under multilingual instructions, which further highlights the challenges of multilingual instruction following in embodied tasks. Besides, in case (a)-(c), models fail to understand the instruction and thus perform wrong actions, which is more straightforward and also indicates the importance of improving the language understanding capability of the model under multilingual instructions.

\subsection{Extended Representation Shift Analysis}
\label{subsec:extended-representation-shift-analysis}

\begin{figure}[H]
    \centering
    \includegraphics[width=0.98\linewidth]{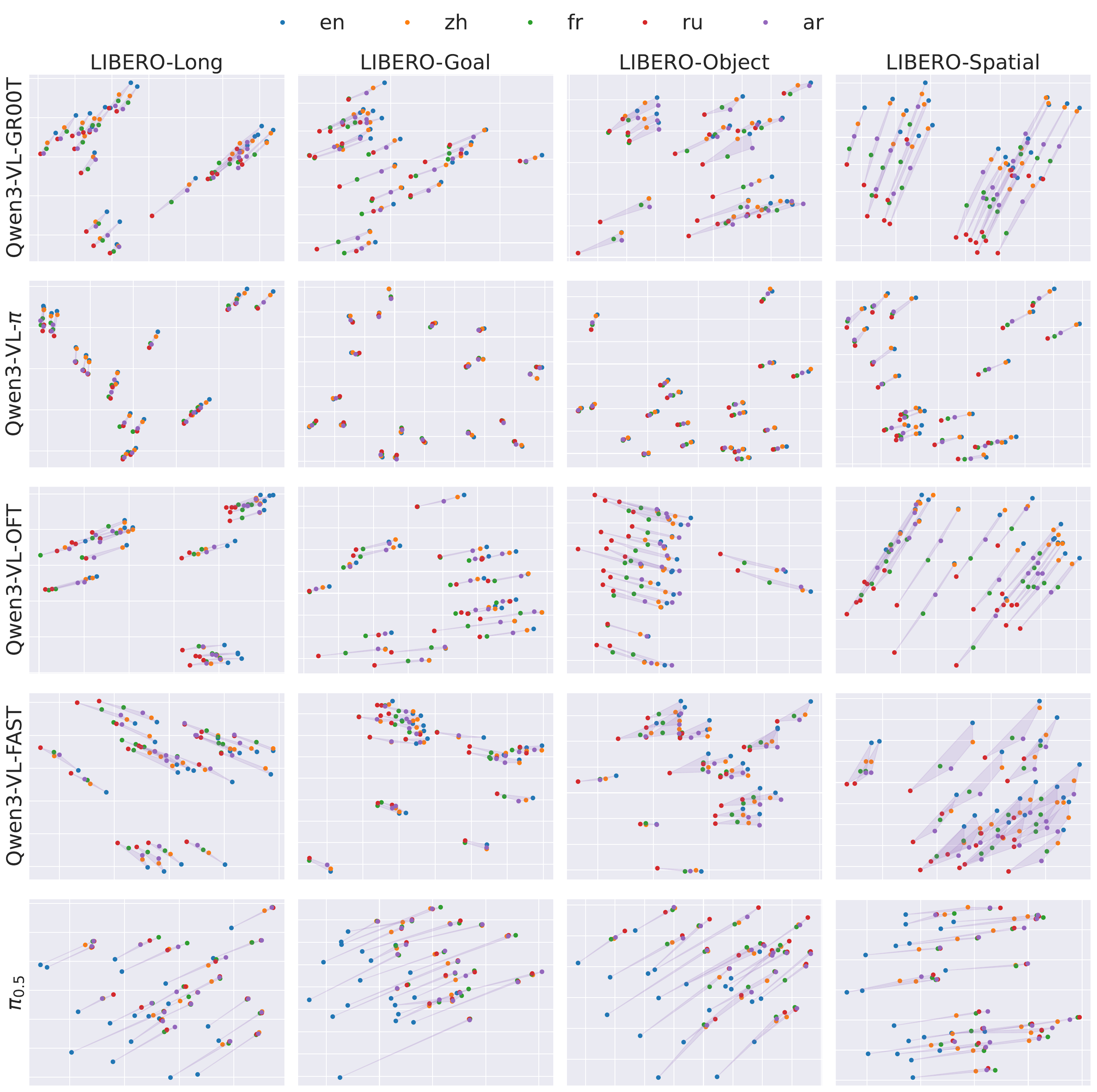}
    \vspace{-0.5em}
    \caption{
        Average pooled embeddings from the middle layer of the model under English instruction and multilingual instruction. The representation shift is observed across multiple models and suites, which indicates that the representation shift under multilingual instructions is a common phenomenon across different models and different suites.
    }
    \label{fig:all-models-all-datasets-embeddings}
\end{figure}

We provide additional visualization of the middle-layer representation shift across all suites in LIBERO, as shown in Figure~\ref{fig:all-models-all-datasets-embeddings}. The results show that the representation shift under multilingual instructions is a common phenomenon across different models and different suites, which further confirms our previous analysis in Section~\ref{subsubsec:representation-shift-analysis}.

\subsection{Extended Ablation Study of MPCA}
\label{subsec:extended-ablation-study-mpca}

\begin{table}[H]
\caption{Ablation study of MPCA. Multilingual performance of different variants of MPCA on LIBERO across four suites. Absolute performance is shown. \textbf{Bold} denotes the best performance under each language. The Avg. denotes the average performance under each language across the four suites. Percentage sign is omitted for better readability.}
\centering
\small
\setlength{\tabcolsep}{3.5pt}
\renewcommand{\arraystretch}{1.15}
\resizebox{\textwidth}{!}
{
\begin{tabular}{c|ccccc|ccccc|ccccc|ccccc|ccccc}
\toprule

\multirow{2}{*}{Models}
& \multicolumn{5}{c|}{Long}
& \multicolumn{5}{c|}{Goal}
& \multicolumn{5}{c|}{Object}
& \multicolumn{5}{c|}{Spatial}
& \multicolumn{5}{c}{Avg.} \\
\cmidrule(lr){2-6} \cmidrule(lr){7-11} \cmidrule(lr){12-16} \cmidrule(lr){17-21} \cmidrule(lr){22-26} 

\multicolumn{1}{c|}{}
& en & zh & fr & ru & ar
& en & zh & fr & ru & ar
& en & zh & fr & ru & ar
& en & zh & fr & ru & ar
& en & zh & fr & ru & ar \\

\midrule
w/o PCA
& 82.5 & 85.5 & \textbf{86.5} & 82.0 & 79.0
& 93.5 & 93.5 & \textbf{79.5} & 69.0 & 87.5
& 97.0 & 87.5 & \textbf{97.5} & 87.0 & \textbf{97.5}
& 95.5 & 85.0 & \textbf{97.0} & \textbf{91.0} & \textbf{93.5}
& 92.1 & 87.9 & \textbf{90.1} & 82.3 & 89.4 \\

MPCA
& \textbf{90.0} & \textbf{90.0} & \textbf{86.5} & \textbf{85.5} & \textbf{83.5} 
& \textbf{97.0} & \textbf{95.5} & 77.5 & \textbf{80.0} & \textbf{92.0} 
& \textbf{98.0} & \textbf{88.0} & 97.0 & \textbf{95.0} & \textbf{97.5} 
& \textbf{96.0} & \textbf{89.0} & \textbf{97.0} & 81.5 & 90.0 
& \textbf{95.3} & \textbf{90.6} & 89.5 & \textbf{85.5} & \textbf{90.8} \\

\bottomrule
\end{tabular}
}
\label{tab:ablation-study-mpca}
\end{table}

We provide an ablation study of MPCA in Table~\ref{tab:ablation-study-mpca} to further analyze the contribution of the component in MPCA. The variant, w/o PCA, denotes the variant without the PCA projection, which directly computes the cosine similarity in the original feature space. The results show that the PCA projection in MPCA can help improve the performance under multilingual instructions. Since the principal components obtained by PCA can capture the most important variance in the data, projecting the features onto these components and aligning them in the subspace can help the optimization process focus on the most important features and thus improve the performance under multilingual instructions, instead of treating all features equally in the original feature space. The results further confirm the effectiveness of MPCA in improving the multilingual performance by addressing the representation shift under multilingual instructions.

\end{document}